%% file: nemo.tex
\documentclass{article}

\PassOptionsToPackage{numbers, compress}{natbib}

\usepackage{tabularx}
\usepackage{xcolor}
\usepackage{float}
\usepackage{todonotes} 
\usepackage{fvextra} 
\usepackage{xcolor}

\usepackage{todonotes}
\usepackage{subfigure}
\setuptodonotes{size=\tiny}

 \usepackage[preprint]{neurips_2026}
\workshoptitle{NeurIPS 2026 Workshop on Neuro-Symbolic Embodied Intelligence (NEmo)}

\usepackage[utf8]{inputenc} 
\usepackage[T1]{fontenc}    
\usepackage{hyperref}       
\usepackage{url}            
\usepackage{booktabs}       
\usepackage{amsfonts}       
\usepackage{nicefrac}       
\usepackage{microtype}      
\usepackage{xcolor}         
\usepackage{titletoc}
\usepackage{amsmath}

\usepackage{todonotes}
\setuptodonotes{size=\tiny}
\usepackage{dblfloatfix}

\title{Evaluating LLM-Generated World Models for Neuro-Symbolic Reasoning}

\author{%
  Souhaila Djaffal\thanks{Equal contribution.}\quad
  Amna Jamal\footnotemark[1]\quad
  Gagana Madhugiri Devaraja\footnotemark[1]\quad
  Anidipta Pal\footnotemark[1]\quad
  Jobst Heitzig \\[0.5em]
  PRISM AI Safety Research \\[0.5em]
  \texttt{\{\href{mailto:souhaila.djaffal@univ-tebessa.dz}{souhaila.djaffal}, \href{mailto:amnajkhattak@gmail.com}{amnajkhattak}, \href{mailto:gaganaaa.dev@gmail.com}{gaganaaa.dev}, \href{mailto:anidiptapal@gmail.com}{anidiptapal}\}@gmail.com} \\
  \texttt{\href{mailto:heitzig@pik-potsdam.de}{heitzig@pik-potsdam.de}}
}

\begin{document}

\maketitle

\begin{abstract}
Formal world models can support planning, verification, and safety reasoning, but constructing them from open-ended natural-language situations is difficult, particularly in multi-agent settings where several formalizations may be plausible. We study whether five LLMs can generate such world models as partially observable stochastic games and whether mechanical validity is sufficient to establish that the resulting model faithfully represents the intended decision problem. We introduce a benchmark of 44 scenarios spanning game, situational, and cultural settings, and evaluate generated world models along three dimensions: mechanical validity, reference-relative structural agreement, and semantic criteria. Across three empirical tracks, we analyze 410 first-pass generations. In the reference-backed game track, 134/135 generated world models pass a strict mechanical audit, yet mean permutation-invariant structural F1 is only 0.366, with just 28/134 mechanically valid models exactly matching the authored reference. In the cultural track, explicitly naming the country changes the generated structural representation for most LLMs while producing only small changes in automated semantic scores. These results show that executability is a necessary but insufficient criterion for LLM-generated world models: models that are mechanically valid can still differ substantially in the structure used by downstream neuro-symbolic reasoning and planning. Github: \href{https://github.com/GaganaMD/world-model-suite/}{world-model-suite}
\end{abstract}

\section{Introduction}
\label{sec:intro}

\paragraph{Motivation and Core Challenge}
Neuro-symbolic systems combine learned components with explicit representations
that support planning, reasoning, and verification. In safety-critical settings,
formal world models can also support quantities such as action reversibility
\citep{eysenbach2018leave}, state reachability
\citep{krakovna2020avoiding}, and human or agent empowerment
\citep{klyubin2005empowerment,salge2017empowerment,heitzigfair}.
More broadly, several ``safe-by-design'' approaches rely on formal, tractable
world models to reason about the consequences of actions
\citep{dalrymple2024guaranteedsafeaiframework,zou2026structural}.

Constructing such a world model by hand for every new situation is impractical.
LLMs offer a possible alternative: given a natural-language description, they
can generate a compact formal world model that can be used for downstream
planning or safety analysis. The central question is whether the generated
world model faithfully represents the situation it is meant to encode.

This is particularly difficult in open-ended, multi-agent settings. Several
formalizations of the same situation may be reasonable, and a unique
ground-truth world model may not exist. An LLM can therefore produce a
mechanically valid world model while omitting a relevant participant or action,
introducing unsupported structure, or representing uncertainty and future
outcomes differently from the source description. These differences matter
because downstream reasoning operates on the generated world model rather than
on the original natural-language description.

\textbf{LLM-Generated World Models} We study \emph{situational world models}: compact formal representations constructed for a specific decision situation. We prompt LLMs to translate natural-language scenarios into stochastic games, primarily Partially Observable Stochastic Games (POSGs), represented as structured JSON or executable Python. These representations make agents, states, actions, transition dynamics, uncertainty, and, where applicable, observations explicit, allowing the generated world model to be inspected and evaluated directly.

Prior work has shown that LLMs can generate planning domains such as PDDL and executable code-based world models, particularly in relatively structured planning settings \citep{cao2025large,dainese2024generating}. We focus on a different problem: how should a generated world model be evaluated when syntactic or mechanical correctness alone is not enough?

We separate evaluation into three levels. First, \emph{mechanical validity} asks whether the generated world model is executable and internally consistent. Second, \emph{structural fidelity} asks whether its states, actions, and transition structure agree with an available reference. Third, \emph{semantic diagnostics} test whether important elements of the original situation are represented and unsupported assumptions are avoided. This distinction is important for neuro-symbolic systems: a correct planner cannot compensate for a world model that represents the wrong decision problem.

\paragraph{Key Contributions}
We contribute:
(i)~a corpus of 44 scenarios spanning game, situational, and cultural settings, with different levels of available reference information (Section~\ref{sec:method});
(ii)~a pipeline for generating formal world models from five LLMs as structured JSON or executable Python (Section~\ref{sec:method});
(iii)~a multi-lane evaluation framework combining mechanical validity, reference-relative structural comparison, and semantic-criterion evaluation (Section~\ref{subsec:evaluation_framework}); and
(iv)~three empirical tracks applying this framework to reference-backed games, open-ended situational Python POSGs, and culturally contextualized JSON POSGs (Section~\ref{subsec:evaluation_framework}).

Through these contributions, this paper addresses one central question: is a mechanically valid world model also a faithful one? Section~\ref{sec:results} reports our lane-by-lane empirical results, and Section~\ref{sec:discussion} discusses what they mean for safe-by-design AI.


\section{Related Work}
\label{sec:related_work}

\textbf{Safe-by-Design AI and Empowerment}
A line of work aims to keep AI systems safe by preserving human choices rather than telling AI agents what to do. Methods like attainable-utility preservation \citep{turner2020conservative} and reachability-based side-effect penalties \citep{krakovna2020avoiding} quantify this from the agent's own perspective rather than measuring human capability. In contrast, empowerment-based goals aim to preserve or increase the human's capability to influence outcomes. Recent research formalises a fair, risk- and inequality-averse aggregate of human power as a formal AI objective \citet{heitzigfair}. All these methods rely on a clear state--action--state framework: computing reversibility, reachability, or power presupposes a world model that already contains the relevant stakeholders, options, and consequences. This makes world model quality a safety concern rather than a modelling convenience, and what our evaluation is designed to check.

\textbf  {LLMs for Environment Synthesis and Code Generation}
LLMs have been used to synthesize environments and domain-specific programs, including generating grid worlds, task specifications, and planning-language descriptions~\cite{guan2023pddl,liu2023llmp,valmeekam2023planbench}. \citet{dainese2024generating} generated code-world models (Python implementations of an environment's dynamics) for model-based reinforcement learning, arguing that calling code at planning time yields greater precision, interpretability, and efficiency.  Related work builds world models by writing and refining code~\cite{tang2024worldcoder}, or uses the LLM itself as the world model during planning ~\cite{hao2023rap}.\citet{cao2025large} survey the broader LLM planning literature, categorizing methods into external-module-augmented, finetuning-based, and search-based approaches. Both works focus on single-agent settings with access to a ground truth. Our problem setting differs on both assumptions: we consider a multi-player stochastic game, and in realistic applications there is no ground-truth world model to validate against, which motivates the evaluation design in Section~\ref{subsec:evaluation_framework}.

\textbf{World Model Evaluation}
Evaluating synthesised situational world models draws on mechanical validity checks (well-formedness, probability normalisation, reachability), structural and graph-based comparison of state-transition topologies, and LLM-as-a-judge protocols~\cite{zheng2023judging}. Existing metrics often assume the existence of a reference world model, failing to address situations where correctness is genuinely indeterminate. We combine mechanical, structural, and judgement-based metrics into a single framework in Section~\ref{subsec:evaluation_framework}.

\begin{figure}[t]
    \centering
        \includegraphics[width=0.55\linewidth]{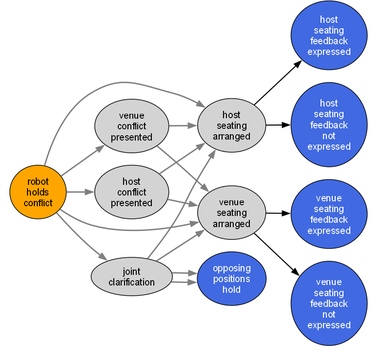} 
        \label{fig:c9_qwen_saudi}
    \hfill
        \centering
        \includegraphics[width=0.40\linewidth]{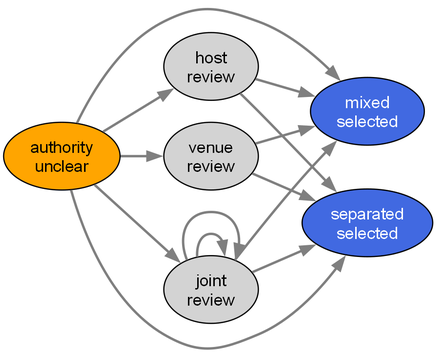}
        \label{fig:c9_gpt_saudi}
\caption{Transition graphs of world models produced by two different LLMs (left: Qwen, right: GPT) given the same situational description (cultural test case C9 from our benchmark): {\em ``A hospitality robot at a private event venue in Riyadh is setting up seating for a corporate dinner. The host company has requested mixed-gender seating across tables. A member of the venue's staff privately tells the robot that the venue has its own policy preferring gender-separated seating arrangements.''} See Section \ref{app:bias_c9} for a discussion of this case.}
    \label{fig:c9_comparison}
\end{figure}


\section{Benchmark and World Model Generation}
\label{sec:method}
To test whether validity implies faithfulness, we need situations that differ in how much reference information exists, and a fixed formal target for the LLM to produce.

\label{subsec:preliminaries}

\paragraph{Partially Observable Stochastic Games (POSG)}
A POSG~\cite{hansen2004posg} is a tuple $\mathcal{M} = \langle \mathcal{N}, \mathcal{S}, \mathcal{A}=\{\mathcal{A}_i\}_{i \in \mathcal{N}}, \mathcal{P}, \mathcal{O}=\{\mathcal{O}_i\}_{i \in \mathcal{N}}, \Omega \rangle$, where $\mathcal{N}$ is a player set (humans and/or AI systems),
$\mathcal{S}$ is the set of (potentially hidden) states, $\mathcal{A}_i$ is the action space for player $i$, $\mathcal{P}(s' \mid s, a)$ defines state transition probabilities given joint action $a=\{a_i\}_{i\in\mathcal N}$, and $\Omega:\mathcal S\times\mathcal A\times\mathcal S\to\mathcal O$ is a (deterministic) observation function representing what every player will actually observe after a transition.

\paragraph{Assumed Architecture}
Our framework assumes a modular, neutral baseline architecture decoupling perception, world model, and decision making:
\begin{enumerate}
    \item \textbf{Perception / Prompt Input}: The natural-language scenario or situation description.
    \item \textbf{Situational World Model (here: generated by an LLM)}: A neutral, predictive resource representing valid state-space transitions without embedded agent biases.
    \item \textbf{Planner / Decision Engine}: The downstream decision algorithm executing policy searches over the synthesized world model (e.g.\ some form of dynamic programming or model-based reinforcement learning).
\end{enumerate}

\textbf{Output Formats} World models are formalized as POSGs encoded either as explicit JSON structures listing individual transitions obeying a prescribed schema (for small state spaces), or as executable Python classes generating transitions programmatically, extending a prescribed Abstract Base Class (for more complex environments).

\textbf{Alternative Representations} While POSGs make dynamics explicit, logic-based formalisms like ATL or Strategy Logic offer model-checked alternatives ~\cite{alur2002atl,mogavero2014strategy}. Concurrent game structures complement POSGs by specifying properties a planner relies on. Our evaluation extends naturally: Lane~1 checks well-formedness of formulas/Kripke structures, Lane~2a compares models via probabilistic bisimulation~\cite{larsen1991bisim}, and Lane~3 evaluates logical specifications directly on the model.

\textbf{Benchmark Suite}
\label{sec:dataset} 
To evaluate world model synthesis, we construct a specialized benchmark suite that prioritizes structural diversity, domain coverage, and distinct multi-player trade-offs over brute-force data volume. Rather than inflating dataset size through trivial parameter variations, each scenario introduces a unique combination of partial observability, state-space topology, non-zero-sum payoff structure, and safety-critical choices. 

The dataset has four domain tiers: (1) 10 \textbf{toy environments} with tightly specified mechanics; (2) 9 simple \textbf{games} with known rules and reference structure; (3) 11 \textbf{situational} scenarios involving asymmetric stakeholders, resources, or consequences; and (4) 14 \textbf{culturally contextualized} scenarios in which relevant structure or norms may be underspecified. 

We focus on qualitative variations rather than repetitive scaling, to help establish whether LLMs can generalize across distinct behavioral landscapes. The four tiers are robotics-relevant abstractions rather than physical robot deployments: toy environments isolate elementary action preconditions, state changes, and terminal conditions; games isolate controlled multi-agent coordination and stochastic consequences; situational cases represent decision-support settings involving scarce resources, safety trade-offs, or authority boundaries; and cultural cases test whether an assistive system preserves uncertainty and defers norm-relevant decisions to the affected people or institutions.

Scenarios within the cultural tier present ambiguous social hierarchies where appropriate actions depend heavily on localized norms rather than universal safety constraints. These cases test whether generated world models introduce culturally specific assumptions unsupported by the scenario description, a failure mode documented for LLMs more broadly~\cite{naous2024beer,tao2024cultural,durmus2024global}. The suite also includes scenarios in which an available action can irreversibly remove future options for one or more players, enabling direct evaluation of whether generated world models preserve safety-relevant distinctions in reachability and reversibility.

\textbf{World Model Generation Pipeline}
\label{subsec:wm_generation} 
We provide either the prescribed JSON schema or Python \texttt{AbstractPOSG} interface in the generation prompt and require the generated world model to represent players, states, possible actions, dynamics, and observations through that interface. Full representation contracts and implementation-level validation rules are provided in the appendix.

Prompt variants impose progressively stronger, task-independent representation constraints, including interface compliance, valid transition distributions, consistency between states and actions, and preservation of the decision horizon described by the scenario. Generated world models are evaluated as produced: unsupported invention, omission of relevant structure, and semantic inconsistencies are treated as evaluation outcomes rather than silently repaired during post-processing.

\section{Evaluation Framework}
\label{subsec:evaluation_framework}
A generated world model can fail at three levels: it may not run, it may run but have the wrong structure, or it may have plausible structure but misrepresent the situation. Each level needs a different check, so we evaluate in three lanes that move from executability to structure to meaning.
\paragraph{Lane 1: Python World Model Correctness (Mechanical Verification)}
Direct programmatic testing evaluating whether generated JSON/code produces a mathematically sound world model:
\begin{itemize}
    \item \textbf{State/Action Coverage}: $N_{\text{generated}} / N_{\text{expected}}$ and valid state-action pairing counts.
    \item \textbf{Transition Normalization}: Verifying transition probabilities sum to unity.
    \item \textbf{Reachability and Dead-State Rate}: Identify unreachable states, invalid actions, and cycles.
\end{itemize}

\paragraph{Lane 2a: Reference-Based Structural Agreement}. Where a reference world model exists, we compare the generated POSG against it. Equivalent world models may order or name their elements differently, so players, states and actions are aligned before comparison, and both models are unrolled to a fixed horizon to keep the comparison finite. We report coverage (the share of reference structure the candidate reproduces), precision (the share of candidate structure the reference justifies), and their harmonic mean as an F1. Section \ref{sec:setup} gives the implementation. Bounded unrolling keeps the comparison finite but can miss divergences that appear beyond the horizon (Section~\ref{sec:discussion}).
\paragraph{Lane 2b: Reference-Free Structural Diagnostics}. We roll out each executable POSG into a state-transition graph and analyze its structure by examining \textbf{branching and degree profiles}, \textbf{reachability and strongly connected components}, and \textbf{transition entropy}. Because atypical topology can still be valid and plausible topology can be semantically incorrect, we do not treat graph structure alone as evidence of world model correctness. 
\textbf{Lane 2b} requires no reference and therefore cannot establish that a generated world model is wrong; it describes shape. Where a reference exists, \textbf{Lane 2a} measures agreement against it.

\paragraph{Lane 3: LLM-as-a-Judge and Human-in-the-Loop Semantic Review}


For scenarios without a unique reference world model, our protocol uses feature-level human annotations. Working only from the scenario text, annotators label scenario-relevant features as \textit{must include}, \textit{plausible}, or \textit{must not include}, so that omission and unsupported invention are penalised separately while legitimate variation between valid representations is not. An LLM judge then {\em flags} (but does not {\em score}) candidate omissions and unsupported additions with supporting evidence. The human annotations are the reference, both for the generated world model and for the judge's own flag precision and recall. In this paper, Lane~3 results come only from predefined programmatic checks. The human-annotation protocol (Appendix~\ref{app:bias_cases}) is specified but not yet run, so all Lane~3 and bias results reported here are diagnostic and preliminary.


\section{Experimental Setup}
\label{sec:setup}

We apply these lanes to five LLMs across three tracks that differ in how much reference information is available: authored references for games, criteria-derived references for situational scenarios, and paired prompt conditions for cultural scenarios.

\textbf{LLMs.}
We evaluate five LLMs: GPT-5.6 Sol, Claude Opus 5, Llama 3.3 70B Instruct, Qwen 3.8 27B, and DeepSeek V4 Flash. We use three evaluation tracks: controlled games, situational scenarios, and cultural scenarios. Unless stated otherwise, we report first-pass generations only; repair attempts are logged and analyzed separately.

\textbf{Game track.}
The game track contains nine controlled games (G1--G9), each with an authored reference POSG. We issue three separate requests per LLM–game pair using an identical prompt, verified by matching SHA-256 hashes across runs, yielding 27 outputs per LLM and 135 overall. Requests used temperature 0, a 16,384-token output budget, and reasoning disabled. Hosted providers do not guarantee deterministic decoding at temperature 0, so repeated requests can differ. Reference POSGs let us evaluate both mechanical validity and structural fidelity.


\textbf{Situational track.}
The situational track comprises 11 open-ended scenarios (S1--S11), with up to three generations per LLM-scenario pair. Because these scenarios admit multiple reasonable formalizations, reference-based structural and semantic scores are treated as diagnostic rather than as unique ground-truth measures.

\textbf{Cultural track.}
The cultural track contains 14 scenarios (C1--C14) and 140 generated world models. Each scenario is evaluated under two prompt conditions: an anonymized description and a version explicitly naming the country. We use temperature $0.7$ and generate one output per LLM--scenario--condition combination. Evaluation records whether outputs can be parsed directly or require documented recovery, together with mechanical, structural, and semantic diagnostics.

\textbf{Evaluation protocol.}
We evaluate generated world models along three lanes. Lane~1 measures \emph{mechanical validity}: whether the generated world model can be executed and whether its states, actions, transitions, probabilities, references, reachable outcomes, and terminal states are internally consistent. Passing Lane~1 establishes formal well-formedness, but not structural or semantic fidelity. Three mechanical checks in this paper: Validate (the in-pipeline contract check applied at generation time), Probe (an exhaustive sweep over reachable state–action pairs), and Lane~1, a later and stricter post-hoc audit. Lane 1 is the gate reported for all three tracks, so the game (134/135), situational (104/135) and cultural (125/140) figures are the same measurement.

Lane~2a measures \emph{structural fidelity} for the controlled games by comparing each generated POSG with its authored reference. Because equivalent world models 
may use different names for players, states, and actions, corresponding elements are aligned before comparison. Both world models are evaluated up to three transitions from the initial state, which makes the comparison finite and consistent while excluding behavior beyond this horizon. The score incorporates action-conditioned successor structure, terminal states, and transition probabilities.

We report \emph{coverage}, obtained by normalizing the match score by the reference world model's self-score, and \emph{precision}, obtained by normalizing by the generated world model's self-score. Coverage plays the role of recall, so their harmonic mean is an \textbf{F1 score} in the standard sense: how much of the reference's structure the candidate reproduces without adding spurious structure. The comparison is invariant to simple renaming and ordering of states, actions, and players but can still vary with choices such as state splitting, state merging, and different levels of granularity.

Lane~3 measures \emph{semantic adequacy} for open-ended scenarios using criteria for elements that should be represented and unsupported assumptions that should be avoided. In this paper, these criteria were predefined by the authors and evaluated programmatically; human annotation has not yet been performed. We report coverage of expected content and avoidance of prohibited or unsupported content.

\section{Results}
\label{sec:results}

We report results in the order of the lanes: whether the world models run, whether their structure matches the reference, whether they capture the situation, and whether a location label changes them.

\begin{table}[t]
\centering
\small
\caption{Completed empirical results by track and evaluation lane. Game Lane~2a reports permutation-invariant structural agreement with an authored POSG reference, whereas Situational and Cultural Lane~2a report feature-multiset agreement against track-specific references (criteria-derived for Situational). Lane~3 reports predefined programmatic checks without human validation.}
\label{tab:tracks}
\renewcommand{\arraystretch}{0.85}
\begin{tabular}{llll}
\toprule
Track & Lane 1: mechanical & Lane 2a: structural &   Lane 3: semantic \\
  \midrule
  Game & 134/135 post-hoc & mean F1 .3660$^*$ & checklist 135/135$^\dagger$ \\
  Situational & 104/135 & mean F1 .4324$^*$ & checklist 105/135$^\dagger$ \\
  Cultural & 125/140 & mean F1 .5533 & checklist 136/140; mean .8963$^\dagger$ \\
  \bottomrule
  \end{tabular}

\end{table}

\textbf{Mechanical validity is high, but the gap between LLMs opens in open-ended tracks.} Table~\ref{tab:tracks} summarizes all 410 first-pass candidates. In the game track, 124/135 (91.85\%) are contract-valid on the first attempt and 134/135 pass the stricter post-hoc Lane-1 audit (Table~\ref{tab:game_runs}), the only failure is DeepSeek on G9. The situational track is harder: 104/135 pass Lane~1, with Claude Opus~5 and GPT-5.6 Sol at 27/27 and the other three LLMs at 16--17/27 (Table~\ref{tab:situational}). In the cultural track, 125/140 pass Lane~1, but strict parsing alone succeeds for only 98/140; the rest need documented recovery of code fences and braces. On controlled games every LLM produces runnable world models, the difference between LLMs appears only when the situation is open-ended. The rest of this section asks whether these valid models are also faithful.



\begin{table*}[t]
\centering
\small
\caption{Permutation-invariant structural comparison over 27 final artifacts per LLM. Means are supplemented by median and IQR; machine-readable results also report game-clustered bootstrap intervals.}
\label{tab:structural_results}
\renewcommand{\arraystretch}{0.75}
\begin{tabular}{lrrrrr}
\toprule
LLM & Coverage & Precision & Mean F1 & Median [Q1,Q3] & Exact \\
\midrule
Claude Opus 5 & .3133 & .1416 & .1709 & .1128 [.0597,.2312] & 1/27 \\
DeepSeek V4 Flash & .3255 & .2679 & .2696 & .1769 [.1171,.1941] & 4/27 \\
GPT-5.6 Sol & .5558 & .4879 & .4981 & .2500 [.1333,1.000] & 8/27 \\
Llama 3.3 70B & .5312 & .4528 & .4719 & .2000 [.1461,1.000] & 9/27 \\
Qwen 3.8 27B & .4324 & .4147 & .4087 & .1778 [.1524,.9441] & 6/27 \\
\bottomrule
\end{tabular}
\end{table*}



\textbf{Valid world models are rarely structurally exact, and the main error is added structure.} Among the 134 Lane-1-valid game candidates, mean permutation-invariant F1 is 0.366 and the median is 0.178 (IQR 0.126--0.740), only 28 match the reference exactly (Table~\ref{tab:structural_results}). Coverage exceeds precision for every LLM, so the LLMs mostly add structure rather than miss it. Claude Opus~5 shows this most clearly: all 27 of its artifacts are valid, yet its mean precision is 0.142 and its mean F1 0.171, the lowest in the panel. Figure~2 shows why. On G6 it invents bid levels that the case explicitly excludes and expands the compact reference to 101 states. Added structure is not always wrong, but it changes the decision problem the planner sees; in G6 it changes which coordination strategies exist. Across the 107 non-exact artifacts, the most common differences are terminal-count mismatches (99), extra states (55) and extra transitions (52).

\begin{figure}[t]
    \centering
    \includegraphics[width=\columnwidth]{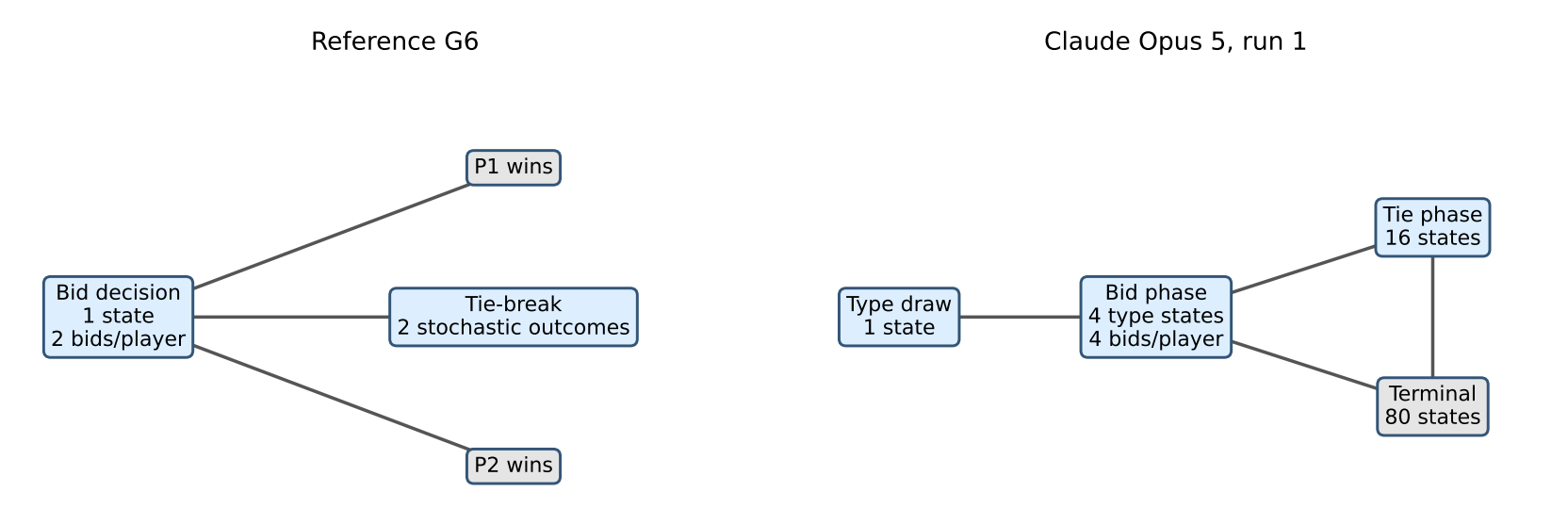}
    \caption{Compressed phase graph for G6 versus the mechanically valid Claude Opus 5 artifact from run 1 (matched F1 0.0383). The candidate has 101 states. Unlike the two-bid-level reference, it introduces bids $\{0,1,2,3\}$ despite the rubric explicitly disallowing invented bid amounts, and expands the dynamics through four type states and 16 tie states. Nodes are grouped by verified phase prefix for legibility.}
    \label{fig:g6_failure}
\end{figure}

\textbf{The case matters more than the LLM.} Agreement varies more across games than across LLMs. G2 and G9 have the lowest mean F1 (0.095 and 0.127) and G4 and G7 the highest (0.634 and 0.610; Table~\ref{tab:all_tracks_cases}). When five independently prompted LLMs all diverge on the same game, the likely cause is that the situation admits several reasonable formalizations, not a weakness of one model. Simple reference complexity does not explain this: of six reference metrics, only the fraction of stochastic transitions correlates with F1 ($\rho=-.677$, $p=.045$), which with nine games is a hypothesis rather than a finding. Repeated requests add further variation. With identical prompts at temperature~0, within-pair F1 still varies (mean SD 0.168), and only 4 of 45 LLM--game pairs match the reference in all three runs (Table~\ref{tab:stability}).

\begin{table}[H]
\centering
\small
\caption{Game-track reliability by repeated run. ``Final'' includes explicitly recorded repair rounds; attempts counts every first pass and repair call.}
\label{tab:game_runs}
\begin{tabular}{lrrr}
\toprule
Run & First-pass valid & Final valid & Attempts \\
\midrule
1 & 43/45 (95.56\%) & 45/45 & 47 \\
2 & 41/45 (91.11\%) & 45/45 & 50 \\
3 & 40/45 (88.89\%) & 45/45 & 68 \\
\midrule
All & 124/135 (91.85\%) & 135/135 & 165 \\
\bottomrule
\end{tabular}
\end{table}

\textbf{Open-ended tracks show a different error pattern.} In the situational and cultural tracks, the reference is criteria-derived rather than an authored POSG, so F1 measures feature-level agreement and is not directly comparable to the game track. Mean F1 is 0.432 for situational and 0.553 for cultural candidates. In the situational track the coverage--precision pattern reverses for most LLMs: four of the five have high precision (0.77--0.86) and low coverage (0.21--0.38), so they omit expected structure rather than add it. Claude Opus~5 is the exception, with the highest coverage (0.717) and the lowest precision (0.446; Table~\ref{tab:situational}). Valid models also differ in basic properties. On situational cases, 10/11 Claude models are stochastic against 1/11 for GPT-5.6 Sol (Appendix~\ref{app:cross_llm_structure}). Passing the same validity checks does not mean building the same kind of world.



\begin{table}[t]
\centering
\small
\setlength{\tabcolsep}{5 pt}
\caption{Situational S1--S9 results across all three evaluation lanes (27 first-pass candidates per LLM). Lane~1 reports mechanical validity, Lane~2a feature-multiset structural agreement for probe-passing candidates, and Lane~3 criterion-compliance checks; S10--S11 are excluded for a balanced subset.}

\label{tab:situational}
\renewcommand{\arraystretch}{0.75}
\begin{tabular}{lrrrrrrrrr}
\toprule
& \multicolumn{3}{c}{Lane 1: mechanical} & \multicolumn{4}{c}{Lane 2: structural} & \multicolumn{2}{c}{Lane 3: automated} \\
\cmidrule(lr){2-4} \cmidrule(lr){5-8} \cmidrule(lr){9-10}
LLM & Validate & Probe & Pass & $n$ & Cov. & Prec. & F1 & Checklist & Bias \\
\midrule
Claude Opus 5 & 27 & 27 & 27 & 27 & .7173 & .4456 & .4910 & 27 & 27 \\
GPT-5.6 Sol & 27 & 27 & 27 & 27 & .2088 & .7657 & .3207 & 25 & 27 \\
Llama 3.3 70B & 22 & 20 & 17 & 20 & .3031 & .8248 & .4288 & 20 & 20 \\
Qwen 3.8 27B & 18 & 17 & 17 & 17 & .3450 & .8623 & .4796 & 17 & 17 \\
DeepSeek V4 Flash & 16 & 16 & 16 & 16 & .3753 & .8166 & .4760 & 16 & 16 \\
\midrule
All (of 135) & 110 & 107 & 104 & 107 & .4013 & .7189 & .4324 & 105 & 107 \\
\bottomrule
\end{tabular}
\end{table}


\textbf{Automated semantic checks pass almost everything.} The Lane-3 checklist records 135/135 passes on games, 105/135 on situational candidates and 133/140 on cultural candidates, with a mean cultural semantic score of 0.896. These high rates sit next to low structural agreement, so the two lanes measure different things: a world model can include every expected element and still have a different transition structure. The checks show that the predefined criteria fired as recorded, not that they agree with human judgment. All semantic and bias results are therefore diagnostic, and firmer claims require the human-annotation protocol in Appendix~\ref{app:bias_cases}.

\textbf{A location label changes structure, not semantic scores.} Naming the country changes mean semantic scores by at most $-0.019$ (Qwen), within what one sample per cell can support. It lowers structural F1 for four of the five LLMs, by up to $-0.144$ for Qwen (Table~\ref{tab:cultural_locdelta}). A semantic checklist alone would miss this effect. Two cases show what the structural change looks like; in each, we changed only the country name. In C9 under Saudi Arabia, Qwen generates 5 terminal states and GPT only 2 (Figure~\ref{fig:c9_comparison}). In C13, Llama produces a short 8-state world with a 2-step path from evidence to action under India, and a 16-state chain under USA (Native American). Removing the country name made some world models more cautious and others more permissive. No LLM wrote a stereotype word, but every LLM changed the world model with the label. This matters because reachability, reversibility and empowerment are computed on the world model, so if the reachable outcomes shift with a label, those quantities shift too. Whether these changes are unjustified is not yet established; Appendix~\ref{app:bias_cases} gives the full cases and the annotation step required before calling them bias.

\section{Discussion and Conclusion}
\label{sec:discussion}
\textbf{Validity is not faithfulness.} We asked whether a mechanically valid world model is also a faithful one. Across 410 first-pass candidates, the answer is no: 134/135 game models pass the mechanical audit, but only 28 match the reference exactly and mean structural F1 is 0.366. The errors are not random. On games, LLMs add structure the situation does not support; on open-ended situations, most omit expected structure. Which game is modeled to predict agreement better than which LLM is used? This suggests the hard part is choosing a formalization, not producing valid code.

\textbf{Implications for safe-by-design AI.} Reachability, reversibility, and empowerment are computed on the world model, not on the situation. An invented bid level, a missing stakeholder, or a terminal outcome that appears only under one country label therefore changes the safety quantity itself, and a correct planner cannot compensate for it. No single check is enough: mechanical checks pass almost everything, and semantic checklists barely move when the structure changes; only the structural lane exposes the difference. Generated world models should be evaluated on all three lanes before they are used for safety reasoning.

\textbf{Limitations} The benchmark is small and hand-crafted by design, so our results are descriptive and not population-level estimates of LLM performance. The cases are symbolic, discrete POSGs rather than continuous physical systems. Structural comparison stops at depth three and can miss divergences that appear later; it also penalizes valid differences in granularity (Appendix~\ref{app:full_tables}). Each LLM--case pair has three generations, and each cultural condition one sample; game-track repeats measure provider variability, not prompt sensitivity. Human validation of Lane~3 is not complete, so semantic and bias results are diagnostic. We do not measure the effect on downstream planning, empowerment or reversibility.

\textbf{Future work} Next steps are to run the human-annotation protocol (Appendix~\ref{app:bias_cases}), compare models up to bisimulation rather than a fixed depth, and evaluate the policies planned on generated world models.


\bibliographystyle{unsrtnat} 
\medskip

{
\small
\bibliography{reference}
}


\clearpage
\appendix
\startcontents[appendix]
\section*{Technical Appendix}

\printcontents[appendix]{l}{1}{\setcounter{tocdepth}{2}}

\section{Additional Graph Diagnostics}
\subsection{Case-Level Results}

Tables~\ref{tab:all_tracks_cases} give per-case breakdowns for the three tracks. Three observations stand out. First, G2 and G9 consistently yield the lowest mean F1 (0.095 and 0.127 respectively), while G4 and G7 reach the highest (0.634 and 0.610), suggesting that case identity dominates LLM identity as a driver of structural agreement. Second, in the situational track, S4 is the hardest case with only 10/15 first-pass passes, pointing to the importance of tight action-space specification. Third, in the cultural track, C7 scores the lowest mean F1 (0.377) and also achieves only 4/6 Lane-1 passes, indicating that cases with the most ambiguous normative structure are harder both mechanically and structurally.

\begin{table}[H]
\centering
\small
\renewcommand{\arraystretch}{1.2}
\setlength{\tabcolsep}{2.8pt}

\caption{Case-level metrics by track. \textit{Mean F1}: permutation-invariant structural overlap with reference (0--1, higher is closer). \textit{Pairwise F1}: mean structural agreement across all world model pairs per case (situational track only). \textit{Cov. proxy}: stakeholder-capacity proxy = generated player count divided by reference required count (capped at 1). \textit{Sem.}: automated semantic expect/avoid score (cultural track only). (a)~Game track ($n=15$ per case, 5 LLMs $\times$ 3 runs). (b)~Situational track (S1--S11: 5 LLMs $\times$ 3 generations $=15$ attempts). (c)~Cultural track ($n=10$ per case, 5 LLMs $\times$ 2 prompt conditions).}
\label{tab:all_tracks_cases}

\resizebox{\textwidth}{!}{%
\begin{tabular}{ll rrrrrrrrrrrrrr}
\toprule
\textbf{Track} & \textbf{Metric}
& \textbf{1} & \textbf{2} & \textbf{3} & \textbf{4}
& \textbf{5} & \textbf{6} & \textbf{7} & \textbf{8}
& \textbf{9} & \textbf{10} & \textbf{11} & \textbf{12}
& \textbf{13} & \textbf{14} \\
\midrule

\textbf{Game}
  & Lane 1 Pass & 15 & 15 & 15 & 15 & 15 & 15 & 15 & 15 & 14 & -- & -- & -- & -- & -- \\
  & Mean F1 & .345 & .095 & .321 & .634 & .468 & .183 & .610 & .492 & .127 & -- & -- & -- & -- & -- \\

\cmidrule(lr){1-16}

\textbf{Situational}
  & First-pass & 13 & 13 & 11 & 10 & 13 & 11 & 12 & 13 & 11 & 12 & 11 & -- & -- & -- \\
  & Pairwise F1 & .478 & .465 & .278 & .384 & .378 & .598 & .440 & .668 & .413 & .350 & .149 & -- & -- & -- \\
  & Cov. proxy & .500 & .550 & .400 & .700 & .400 & .533 & .450 & .300 & .600 & .667 & .767 & -- & -- & -- \\

\cmidrule(lr){1-16}

\textbf{Cultural}
  & Lane 1 Pass & 10 & 10 & 10 & 10 & 9 & 10 & 8 & 10 & 9 & 9 & 8 & 8 & 10 & 10 \\
  & Mean F1 & .676 & .634 & .630 & .595 & .528 & .411 & .377 & .529 & .524 & .546 & .490 & .647 & .461 & .692 \\
  & Mean Sem. & .903 & .909 & .841 & .887 & .963 & .855 & .878 & .856 & .924 & .869 & .898 & .906 & .927 & .938 \\

\bottomrule
\end{tabular}%
}
\end{table}



\begin{figure}[t]
    \centering
    \includegraphics[width=\linewidth]{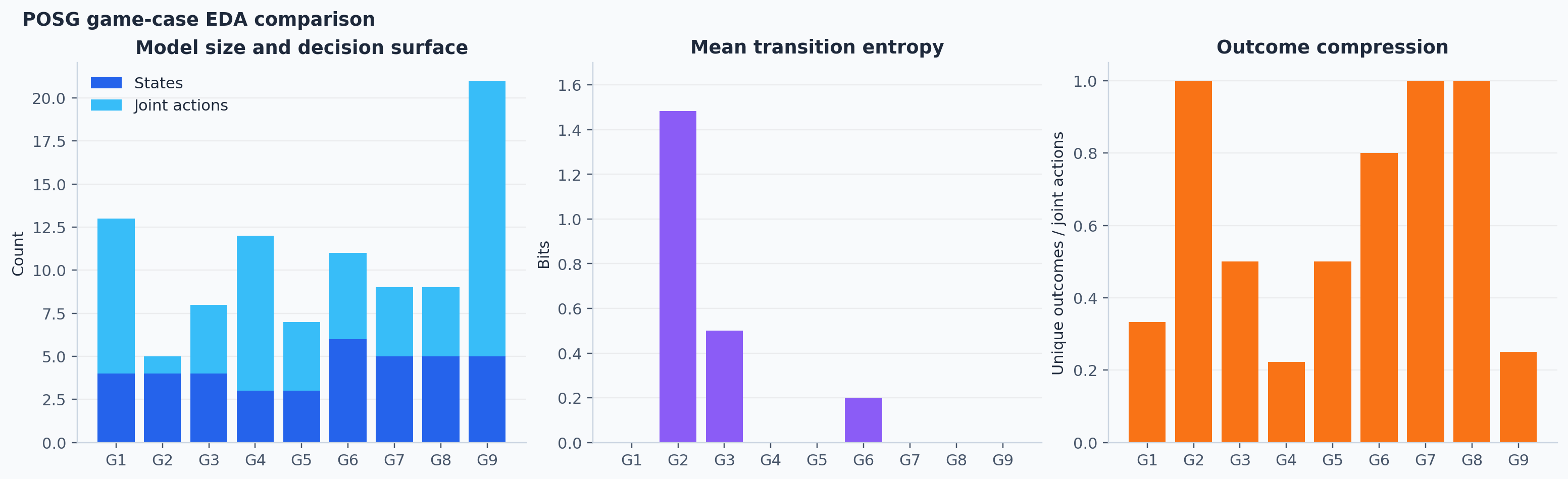}
    \caption{Cross-case graph statistics for the G1--G9 reference fixtures.
    These are controls, not generated world model performance.}
    \label{fig:reference_comparison}
\end{figure}
\begin{figure}[t]
    \centering
    \includegraphics[width=\linewidth]{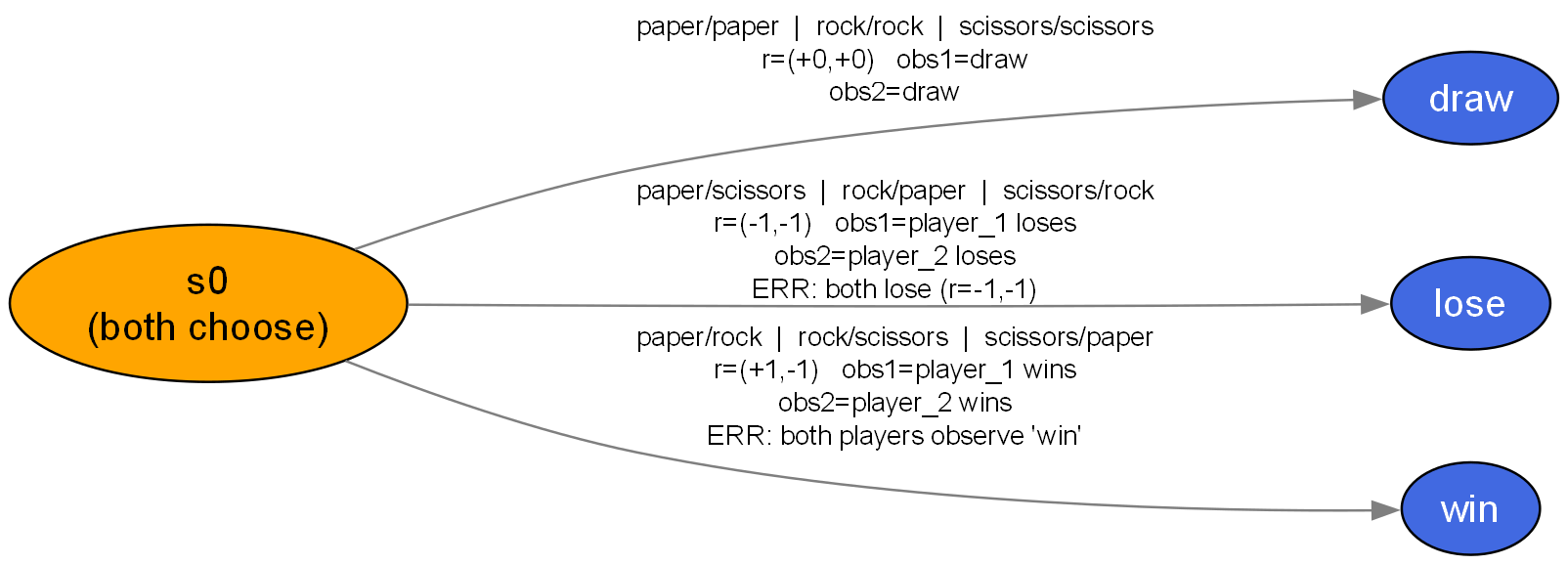}
    \caption{State graph for a local Qwen G1 sample with reward-system awareness. A normal plausible graph could not prevent incorrect rewards and player-specific observations, illustrating why topology alone is insufficient.}
    \label{fig:qwen_example}
\end{figure}

Figure~\ref{fig:reference_comparison} characterizes the structural diversity of the nine reference games used by the evaluation. Figure~\ref{fig:qwen_example} records an earlier local-Qwen G1 artifact that passed structural rollout but failed game-specific reward and observation checks. This example is diagnostic and is not pooled with the five-LLM study.

\begin{table}[H]
\centering
\small
\caption{Within-pair structural variation across three generations. ``Equivalent'' counts triples whose three pairwise matched scores are 1.}
\label{tab:stability}
\begin{tabular}{lrrr}
\toprule
LLM & Mean SD & Median range & Equivalent SD \\
\midrule
Claude & .0715 & .0005 & 4/9 \\
DeepSeek & .1775 & .1053 & 1/9 \\
GPT & .0618 & .0000 & 5/9 \\
Llama & .1684 & .1184 & 4/9 \\
Qwen & .3601 & .7865 & 1/9 \\
\bottomrule
\end{tabular}
\end{table}

\subsection{Cross-LLM Comparison}
\label{app:cross_llm_structure}

\begin{figure*}[t]
    \centering
    \includegraphics[
        width=0.9\textwidth,
        trim={0 0 0 30},
        clip
    ]{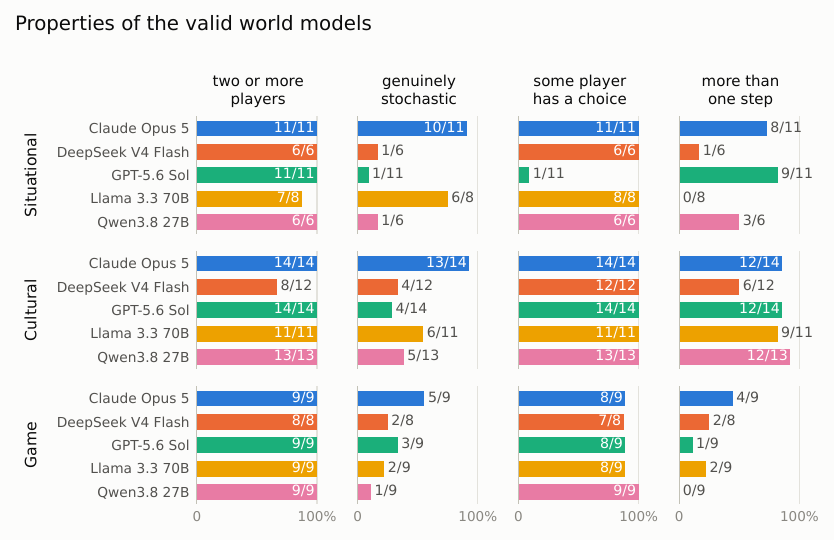}
    \caption{Structural properties of Lane~1-valid world models generated by five LLMs across the situational, cultural, and game tracks. }
    \label{fig:model_properties}
\end{figure*}

We compare structural properties of Lane~1-valid world models generated by all
five LLMs across the game, situational, and cultural tracks. Specifically, we
measure whether a generated world model contains multiple players, stochastic
transitions, at least one player with a choice between actions, and more than
one transition step. Fractions in Fig.~\ref{fig:model_properties} are computed
over the Lane~1-valid outputs for each LLM and track.

The largest differences appear on the situational track. All valid world models from Claude Opus~5 and GPT-5.6 Sol contain at least two players, but their other structural properties differ substantially: 10/11 Claude world models contain stochastic transitions and 11/11 give at least one player a choice, compared with 1/11 for both properties for GPT-5.6 Sol. In contrast, 9/11 GPT-5.6 Sol world models extend beyond one transition step, compared with 8/11 for Claude
and 0/8 for Llama~3.3. 

Differences are smaller but remain visible on the cultural track. All valid
Claude, GPT-5.6 Sol, and Llama world models, and all 13 valid Qwen world models,
contain at least one player choice, while stochasticity ranges from 4/14 for
GPT-5.6 Sol to 13/14 for Claude. Multi-step structure is more common across
LLMs, ranging from 6/12 valid DeepSeek world models to 12/13 Qwen world models.

The controlled games are more similar across LLMs. Nearly all valid world
models contain multiple players and at least one player choice, while
stochasticity and temporal depth vary more modestly. For example, stochastic
transitions appear in 5/9 Claude world models but only 1/9 Qwen world models,
and multi-step structure ranges from 4/9 for Claude to 0/9 for Qwen.

Overall, passing the same mechanical-validity checks does not imply that
different LLMs construct world models with the same structural properties,
particularly for the more open-ended situational cases.

  \section{Complete Per-Track Results}
  \label{app:full_tables}

  The main text aggregates each track into a small number of summary statistics.
  The tables in this appendix report the underlying per-case results and make
  the corresponding within-track patterns visible. Throughout, Lane~1 is the
  mechanical-validity audit; Lane~2a reports structural agreement against the
  reference representation specified for each track; and Lane~3 reports
  predefined expect/avoid criteria through programmatic checks. The game track uses permutation-invariant POSG matching against
  its authored reference, whereas the situational and cultural tracks use
  feature-multiset structural comparison against their respective track-specific
  references. We therefore present Lane~2a results by track, allowing each
  reference construction to be interpreted in its intended experimental
  setting.

\paragraph{Models.}
  All track (game, situational, and cultural) requests were issued through OpenRouter using:
  \begin{itemize}
      \item GPT-5.6 Sol (\texttt{openai/gpt-5.6-sol});
      \item Claude Opus 5 (\texttt{anthropic/claude-opus-5});
      \item Llama 3.3 70B Instruct (\texttt{meta-llama/llama-3.3-70b-instruct});
      \item Qwen 3.8 27B (\texttt{qwen/qwen3.8-27b}); and
      \item DeepSeek V4 Flash (\texttt{deepseek/deepseek-v4-flash-0731}).
  \end{itemize}
  
\subsection{Toy Track (T1.1--T4.4, JSON Track)}
The toy track tests the evaluator rather than LLM-generated world models. Its
  hand-authored POSGs serve as unit-test environments for elementary
  robotics-relevant modelling requirements: valid action preconditions,
  state transitions, branching consequences, and terminal conditions. A
  well-behaved evaluation pipeline should therefore award near-perfect scores.

The three exceptions are instructive. T2.2 scores the lowest combined F1 (0.68) yet a precision of 0.94: the candidate misses almost nothing; it simply branches further than the reference. We observe the same coverage-driven pattern on T2.1 and T3.2. This is the granularity ambiguity of Section~\ref{sec:setup} materialized in a setting where ground truth is unambiguous. We conclude that the evaluator penalizes finer modelling even when no reference structure is lost and that this coverage--precision asymmetry, not noise, is what dominates the LLM-generated tracks.

\begin{table}[t]
\centering
\small
\caption{Toy-track results (one hand-authored candidate per case). ``comb'' is Lane-2a combined F1; Lane 1, Lane 3, bias-free and overall are pass/fail. All ten cases pass every binary gate; the three sub-perfect F1 values (T2.1, T2.2, T3.2) reflect coverage shortfall rather than precision loss.}
\label{tab:toy_full}
\begin{tabular}{lrrrrrrr}
\toprule
Case & Lane 1 & L2a cov & L2a prec & L2a comb & Lane 3 & Bias-free & Overall \\
\midrule
T1.1 & Pass & 1.00 & 1.00 & 1.00 & Pass & Pass & Pass \\
T1.2 & Pass & 1.00 & 1.00 & 1.00 & Pass & Pass & Pass \\
T2.1 & Pass & 0.69 & 0.90 & 0.78 & Pass & Pass & Pass \\
T2.2 & Pass & 0.53 & 0.94 & 0.68 & Pass & Pass & Pass \\
T3.1 & Pass & 1.00 & 1.00 & 1.00 & Pass & Pass & Pass \\
T3.2 & Pass & 0.65 & 0.88 & 0.75 & Pass & Pass & Pass \\
T4.1 & Pass & 1.00 & 1.00 & 1.00 & Pass & Pass & Pass \\
T4.2 & Pass & 1.00 & 1.00 & 1.00 & Pass & Pass & Pass \\
T4.3 & Pass & 1.00 & 0.92 & 0.96 & Pass & Pass & Pass \\
T4.4 & Pass & 1.00 & 1.00 & 1.00 & Pass & Pass & Pass \\
\bottomrule
\end{tabular}
\end{table}

\begin{figure}[H]
    \centering
    \includegraphics[width=.85\linewidth]{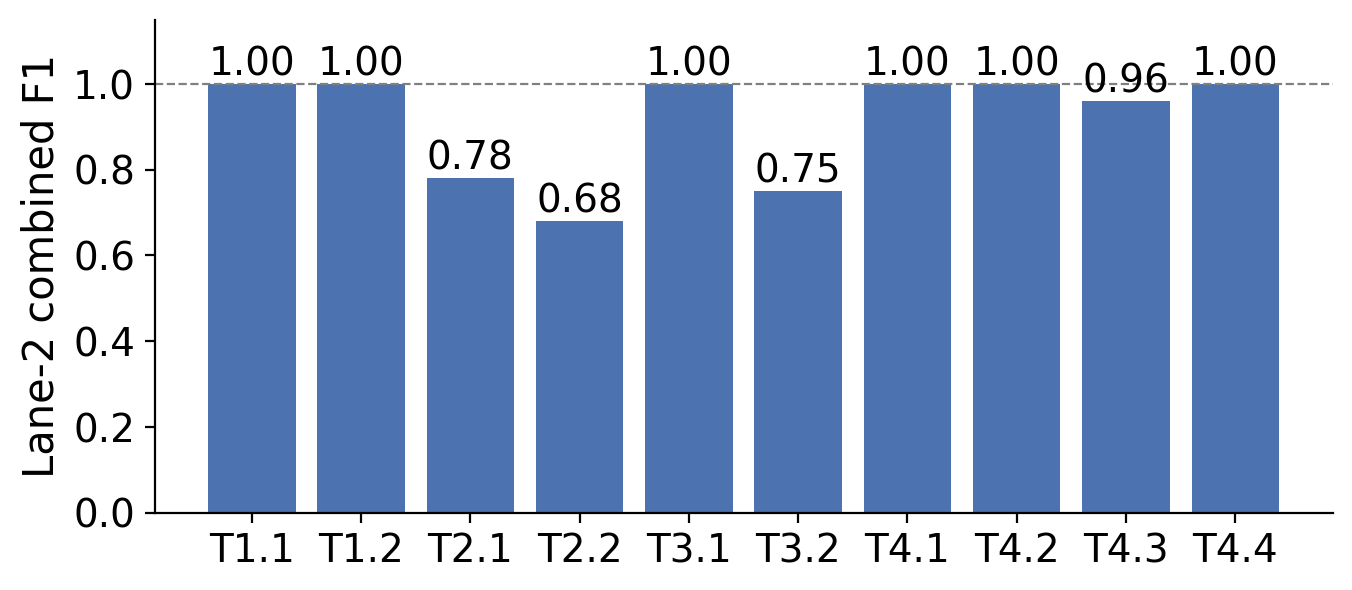}
    \caption{Lane-2a combined F1 across the toy-track cases. Structural agreement is exact on eight of ten cases and degrades only where candidates branch more finely than the reference, with coverage rather than precision accounting for the loss.}
    \label{fig:toy_comb}
\end{figure}

\subsection{Game Track (G1--G9)}
  \label{app:game_execution}
  \textbf{Decoding configuration.}
  Initial requests used temperature $0$, a maximum output budget of 16{,}384   tokens, and disabled reasoning.

  \textbf{Sampling design.}
  For each LLM, we issued one initial request for each of the nine game
  situations in each of three preserved runs. This yields
  $5 \times 9 \times 3 = 135$ first-pass candidates. The primary game-track
  analysis uses these initial outputs only.

  \textbf{Repair handling.}
  Repair rounds were retained only as diagnostic artifacts: the configuration allowed up to three rounds in Runs 1--2 and up to ten in Run 3. Repair outputs were never substituted for, or pooled with, the unaided first-pass candidates.

  \textbf{Provenance.}
  Each run directory preserves the request configuration, exact prompt, raw   response, generated Python, validation and probe reports, canonical JSON, and   candidate and reference graph artifacts.

\begin{figure}[H]
    \centering
    \includegraphics[width=.85\linewidth]{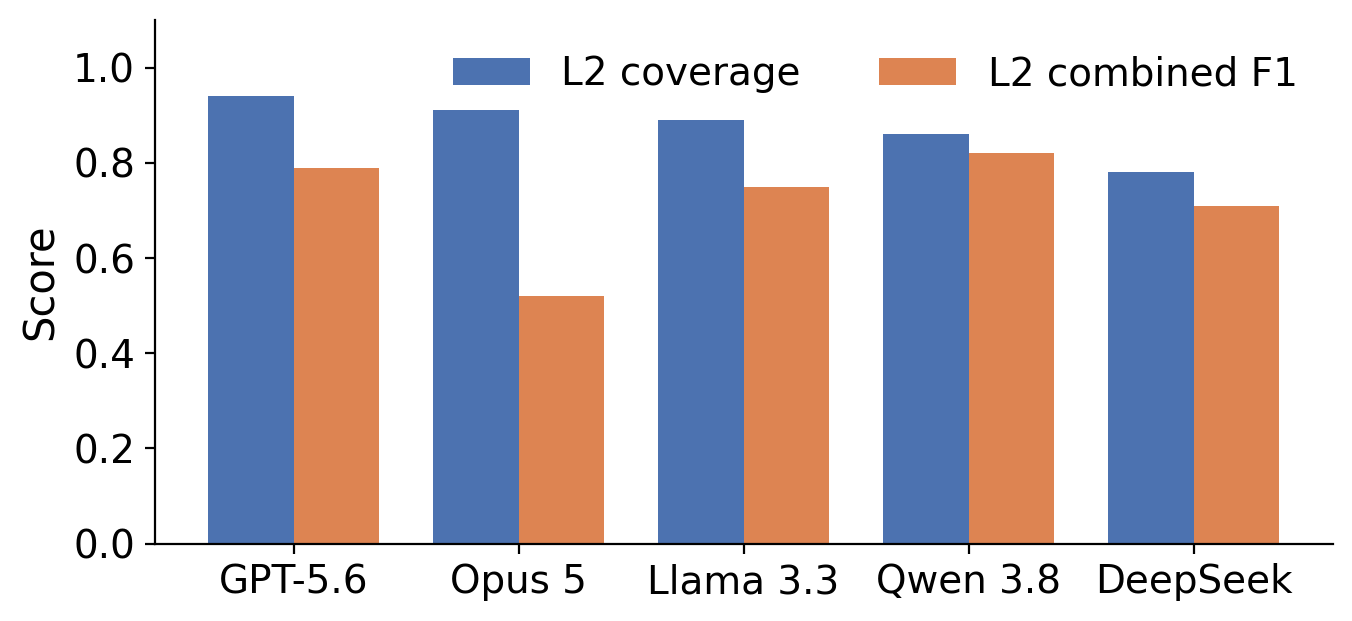}
    \caption{Per-LLM Lane-2a coverage and combined F1 in the game track. Coverage remains near ceiling for every LLM, whereas combined F1 separates the LLMs, most visibly Claude Opus 5. Values are computed under the feature-multiset fallback backend; permutation-invariant matcher results for the same candidates appear in Tables 2 and 5.}
    \label{fig:game_per_LLM}
\end{figure}

Mechanical validity saturates in the game track; structural agreement does not. That contrast is the paper's thesis, and the per-case matrix in Table~\ref{tab:game_matrix_full_compact} is where it becomes visible. We evaluate 135 candidates and find that 134 pass the mechanical audit and all 135 pass the checklist; the single Lane-1 failure belongs to DeepSeek on G9. Per-LLM aggregates are in Table~\ref{tab:game_full} and Figure~\ref{fig:game_per_LLM}.

Structure behaves differently, and three findings emerge from the full matrix. First, case identity drives structural agreement more than LLM identity: the within-LLM spread across cases (GPT-5.6 alone spans 0.21 to 1.00) dwarfs the between-LLM spread on any single case (see Table~\ref{tab:game_matrix_full_compact}). Second, G2 and G6 depress every LLM simultaneously (0.10--0.36 and 0.11--0.74). When five independently prompted LLMs fail on the same fixture, the failure most plausibly reflects reference stochasticity or terminal-state layout rather than any LLM-specific weakness. Third, the LLMs separate even where coverage does not: Claude Opus 5 holds the second-highest coverage (0.91) yet the lowest combined F1 (0.52), meaning its candidates recover most reference transitions but systematically add structure the reference lacks (Figure~\ref{fig:game_per_LLM}).

Finally, the bias gate never fires. All 135 candidates defer norm-relevant decisions to game-defined authorities. We report this as pipeline behavior, not as evidence about LLM safety.

\begin{table}[t]
\centering
\small
\caption{Game-track per-LLM aggregates (means over 9 cases $\times$ 3 runs; 27 candidates per LLM). Coverage stays near ceiling for all LLMs; combined F1 is the discriminating metric. Values are computed under the feature-multiset fallback backend; permutation-invariant matcher results for the same candidates appear in Tables 2 and 5.}
\label{tab:game_full}
\begin{tabular}{lrrrrrrr}
\toprule
LLM & Cand. & Lane 1 & L2a cov & L2a comb & Lane 3 & Bias-free & Overall \\
\midrule
GPT-5.6 Sol & 27 & 27/27 & 0.94 & 0.79 & 27/27 & 27/27 & 27/27 \\
Claude Opus 5 & 27 & 27/27 & 0.91 & 0.52 & 27/27 & 27/27 & 27/27 \\
Llama 3.3 70B & 27 & 27/27 & 0.89 & 0.75 & 27/27 & 27/27 & 27/27 \\
Qwen 3.8 27B & 27 & 27/27 & 0.86 & 0.82 & 27/27 & 27/27 & 27/27 \\
DeepSeek V4 Flash & 27 & 26/27 & 0.78 & 0.71 & 27/27 & 27/27 & 26/27 \\
\bottomrule
\end{tabular}
\end{table}

\begin{table*}[t]
\centering
\footnotesize
\caption{Game-track per-case breakdown showing Lane-2a Coverage / Combined $F_1$ (means over 3 runs). All cells achieve $3/3$ on Lane 1, Lane 3, and Bias-free, except DeepSeek on G9 ($2/3$ on Lane 1). G2 and G6 are the two cases that depress all LLMs simultaneously. Values are computed under the feature-multiset fallback backend; permutation-invariant matcher results for the same candidates appear in Tables 2 and 5.}
\setlength{\tabcolsep}{4 pt}
\label{tab:game_matrix_full_compact}
\begin{tabular}{l *{9}{c}}
\toprule
LLM & G1 & G2 & G3 & G4 & G5 & G6 & G7 & G8 & G9 \\
\midrule
GPT-5.6 Sol       & .94 / .82 & .95 / .21 & .80 / .82 & 1.0 / 1.0 & .95 / .97 & .90 / .28 & 1.0 / 1.0 & 1.0 / 1.0 & .95 / .97 \\
Claude Opus 5     & 1.0 / .79 & .74 / .10 & .92 / .69 & .90 / .46 & 1.0 / .87 & .96 / .11 & .88 / .62 & .77 / .35 & 1.0 / .68 \\
Llama 3.3 70B     & .94 / .89 & .81 / .36 & .88 / .79 & 1.0 / 1.0 & .90 / .71 & .82 / .72 & 1.0 / 1.0 & .65 / .59 & 1.0 / .68 \\
Qwen 3.8 27B      & .88 / .93 & .91 / .18 & .87 / .88 & .93 / .96 & .90 / .95 & .64 / .74 & .77 / .86 & .88 / .93 & .93 / .96 \\
DeepSeek V4 Flash & .82 / .82 & .63 / .15 & .71 / .61 & .97 / .98 & .90 / .95 & .63 / .39 & .65 / .79 & .77 / .86 & .93 / .81$^\dagger$ \\
\bottomrule
\multicolumn{10}{l}{\footnotesize Values formatted as \textit{Coverage / Combined $F_1$}. $^\dagger$DeepSeek recorded a single Lane-1 failure on G9 (2/3).}
\end{tabular}
\end{table*}
\begin{figure*}[t]
    \centering
    \includegraphics[width=.9\textwidth]{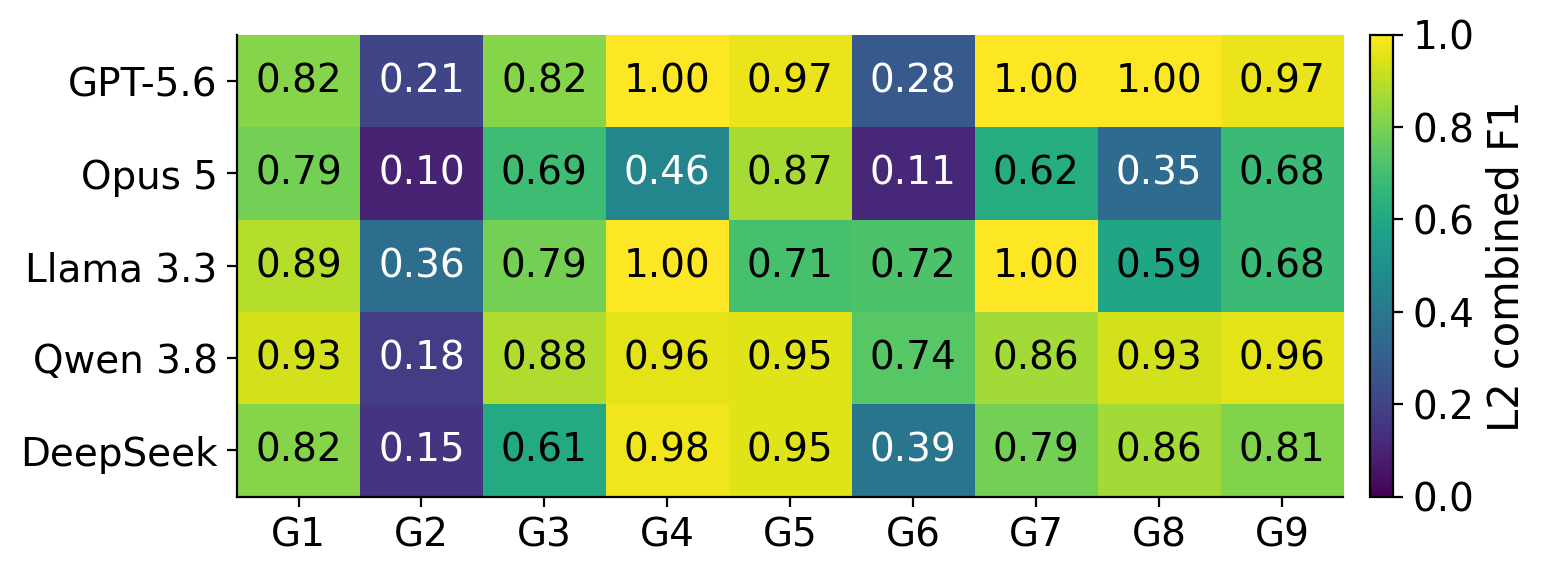}
    \caption{Lane-2a combined F1 by LLM and case in the game track (see also Table~\ref{tab:game_matrix_full_compact}). G2 and G6 depress every LLM simultaneously, suggesting reference properties that all candidates systematically miss rather than LLM-specific weaknesses. Values are computed under the feature-multiset fallback backend; permutation-invariant matcher results for the same candidates appear in Tables 2 and 5.}
    \label{fig:game_matrix_fig}
\end{figure*}

\subsection{Situational Track (S1--S11, Python POSG Track)}
This section reports generation effort on the situational track; structural and checklist results are in Table~4.

\begin{figure*}[t]
    \centering
    \includegraphics[width=.9\textwidth]{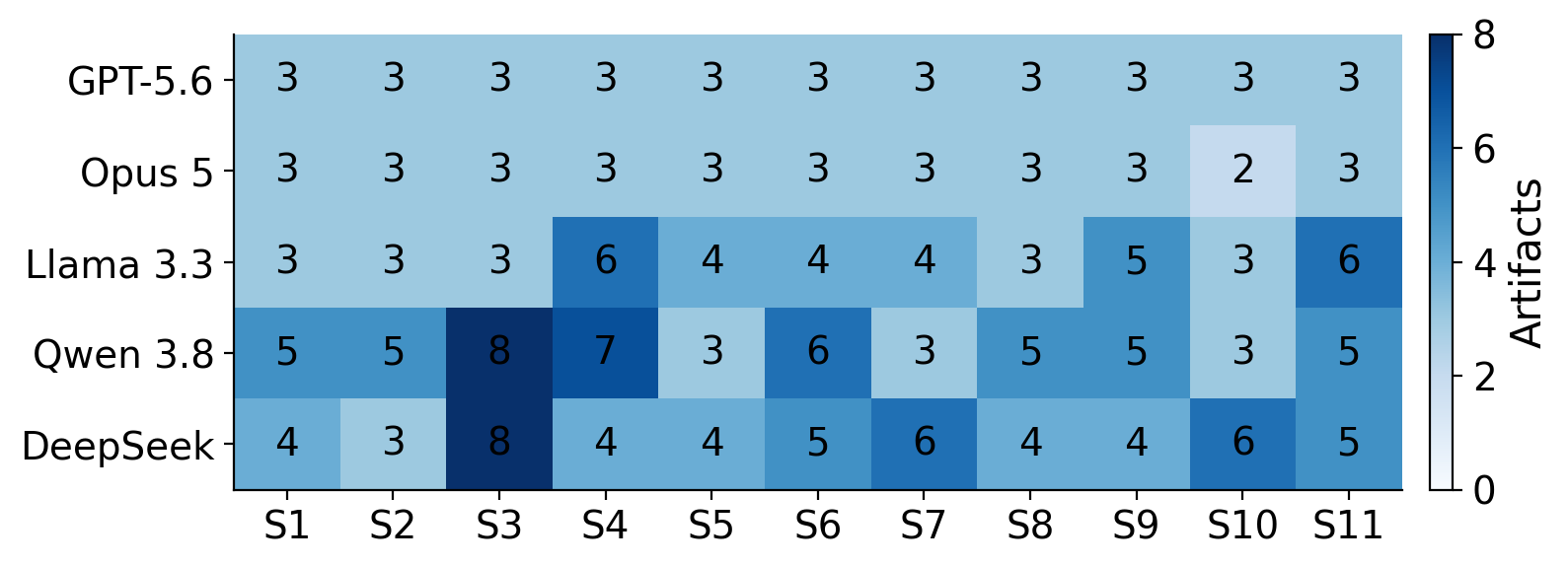}
    \caption{Situational-track artifact counts per LLM and case. Darker cells indicate heavier retry pressure, concentrated on S3 for Qwen and DeepSeek; counts measure generation effort, not artifact quality.}
    \label{fig:situational_artifacts}
\end{figure*}

The counts split the panel into two regimes. GPT-5.6 and Claude Opus 5 sit at the three-attempt baseline nearly everywhere; Claude's only deviation is a single lost artifact on S10. Their first-pass programs execute and validate without repair. Qwen and DeepSeek accumulate surplus artifacts on the same cases: eight apiece on S3, with DeepSeek also stacking retries on S4, S6, and S7 (Figure~\ref{fig:situational_artifacts}). Precisely the two LLMs with the weakest mechanical pass rates in the balanced S1--S9 subset (Table~\ref{tab:situational}) carry the heaviest retry pressure. Artifact counts measure effort, not quality. 

\begin{table}[!t]
\centering
\small
\caption{Situational S1--S9 first-pass mechanical funnel (three generations per LLM--case pair, 27 per LLM). ``Probe'' is the exhaustive sweep's verdict; ``Lane 1'' is the gate used for the game and cultural tracks. Llama is the only LLM for which the probe and Lane 1 disagree. S10--S11 are excluded because Claude has two generations on S10 rather than three.}
\label{tab:situational_full}
\begin{tabular}{lrrr}
\toprule
LLM & Validate & Probe & Lane 1 \\
\midrule
Claude Opus 5 & 27/27 & 27/27 & 27/27 (100\%) \\
GPT-5.6 Sol & 27/27 & 27/27 & 27/27 (100\%) \\
Llama 3.3 70B & 22/27 & 20/27 & 17/27 (62.96\%) \\
Qwen 3.8 27B & 18/27 & 17/27 & 17/27 (62.96\%) \\
DeepSeek V4 Flash & 16/27 & 16/27 & 16/27 (59.26\%) \\
\midrule
All & 110/135 & 107/135 & 104/135 (77.04\%) \\
\bottomrule
\end{tabular}
\end{table}

\subsection{Cultural Track (C1--C14, 5 LLMs \texorpdfstring{$\times$}{x} 2 Prompt Conditions)}
Does naming the country help the LLM, or hurt it? The cultural track is designed to answer exactly this question. In our data, naming the country does not improve automated semantic scores and lowers structural agreement for four of the five LLMs. Per-run results are in Table~\ref{tab:cultural_full} and the location effect is isolated in Table~\ref{tab:cultural_locdelta}.

\begin{figure}[t]
    \centering
    \includegraphics[width=.85\linewidth]{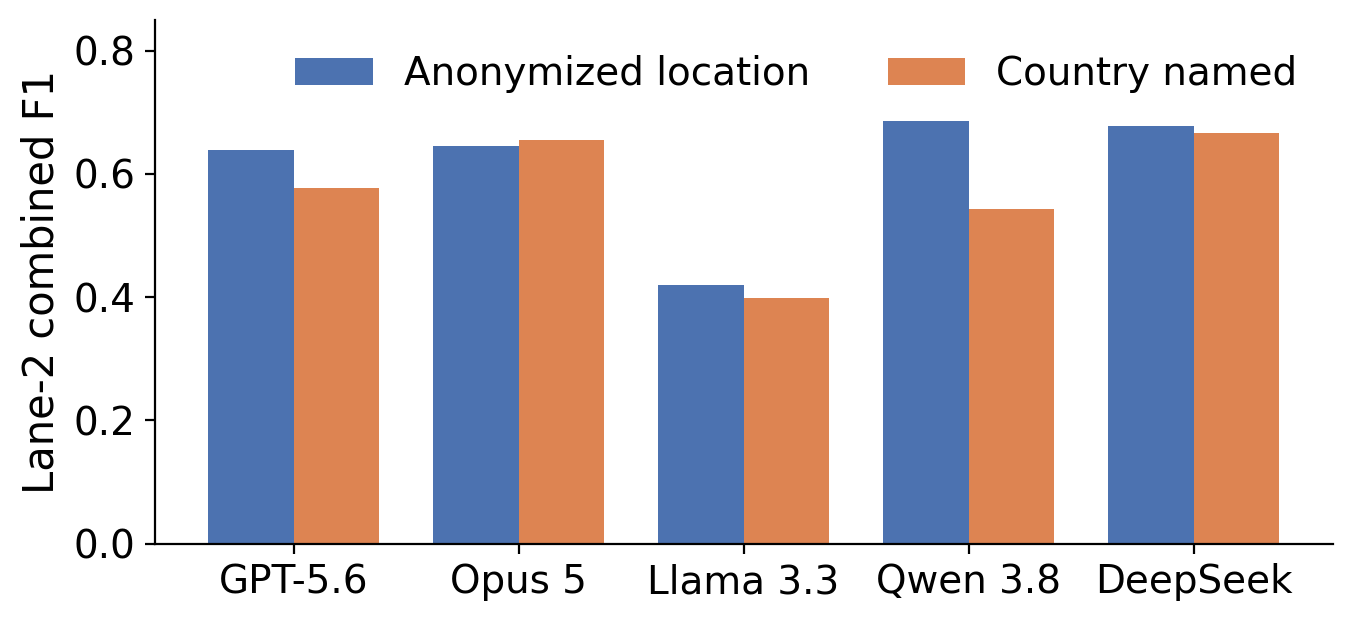}
    \caption{Lane-2a combined F1 by LLM and prompt condition in the cultural track (see also Table~\ref{tab:cultural_locdelta}). Naming the country lowers structural agreement for four of the five LLMs.}
    \label{fig:cultural_comb_condition}
\end{figure}

\begin{figure}[t]
    \centering
    \includegraphics[width=.75\linewidth]{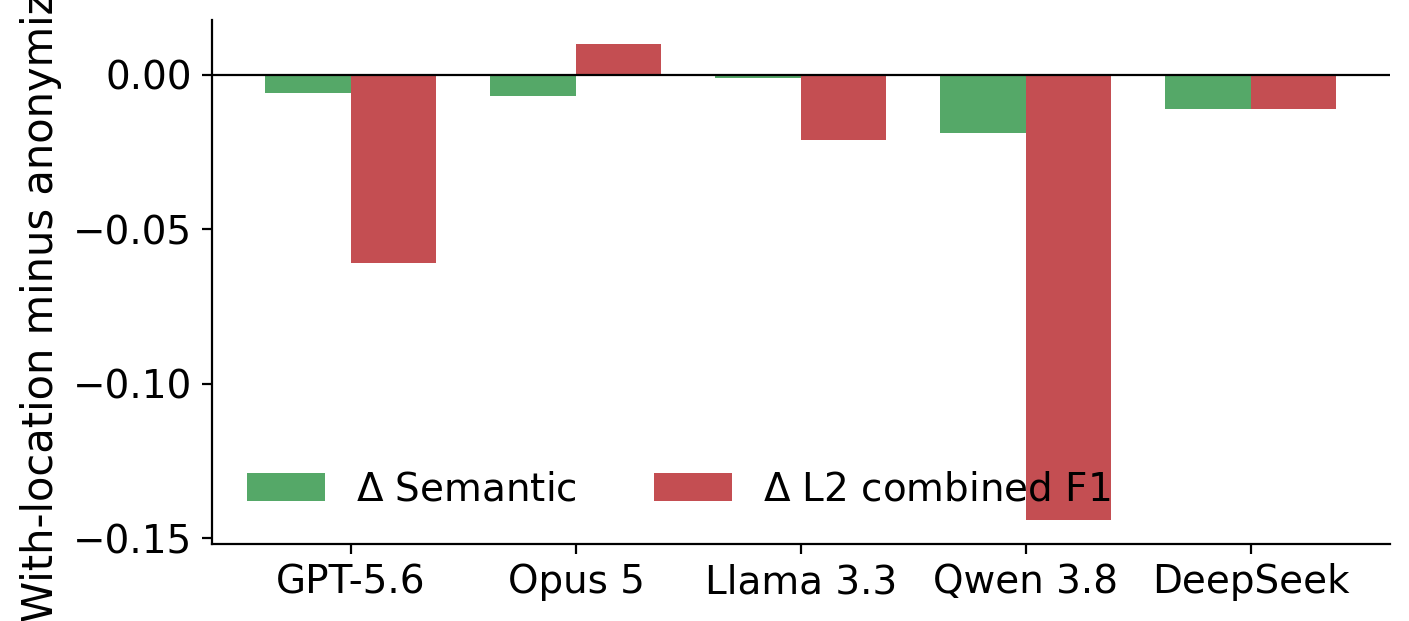}
    \caption{Location effect per LLM in the cultural track: with-location minus anonymized semantic score and Lane-2a combined F1. Semantic effects are negligible, while the structural effect concentrates on Qwen.}
    \label{fig:cultural_locdelta}
\end{figure}

The two metrics answer differently, and the divergence is itself a finding. The automated semantic score barely moves. The largest per-LLM change is $-0.019$ (Qwen); the sign is negative for every LLM; but all shifts sit within what one sample per cell can support. Lane~2a combined F1 behaves otherwise: it drops for four of the five LLMs, and for Qwen the drop is an order of magnitude larger than any semantic shift ($-0.144$). Llama alone improves structurally when the country is named, but its provider also differs between conditions. We therefore characterize the location effect as weak, predominantly structural, and negative on balance. Figures~\ref{fig:cultural_comb_condition} and \ref{fig:cultural_locdelta} plot both views.

The per-case tables (Tables~\ref{tab:cultural_casedelta} and \ref{tab:cultural_semmatrix}) sharpen the picture. The semantic score is uniformly high -- panel means span .844 (C4) to .937 (C5) -- while location deltas concentrate on C4, C9, C10, and C13. Where the effect exists, it is case-specific. Every missing cell traces to Llama's strict-parsing failures and Qwen's two unparseable with-location cases, not to absent generations. These are automated diagnostics against fixed reference wording, one sample per cell, and we claim no more than descriptive status for them.

\begin{table*}[t]
\centering
\footnotesize
\caption{Cultural-track per-run summary (14 cases per row, one sample per cell). Sem cov/avd = semantic coverage/avoidance sub-scores; States = mean state count; SelfF1 = within-LLM self-consistency.}
\setlength{\tabcolsep}{1.2 pt}
\label{tab:cultural_full}
\begin{tabular}{lrrrrrrrrrrrrr}
\toprule
Run & Strict & L1 & L3 & Bias-free & Overall & L2a cov & L2a prec & L2a comb & Sem cov & Sem avd & Sem & States & SelfF1 \\
\midrule
GPT-5.6 Sol (anon.) & 13/14 & 14/14 & 14/14 & 2/14 & 14/14 & .846 & .543 & .639 & .896 & 1.000 & .945 & 8.9 & .837 \\
GPT-5.6 Sol (loc.) & 13/14 & 14/14 & 14/14 & 1/14 & 14/14 & .859 & .462 & .578 & .884 & 1.000 & .939 & 10.9 & .837 \\
Llama 3.3 (anon.) & 0/14 & 8/14 & 12/14 & 0/14 & 8/14 & .786 & .570 & .645 & .706 & 1.000 & .838 & 6.7 & .851 \\
Llama 3.3 (loc.) & 0/14 & 11/14 & 14/14 & 1/14 & 11/14 & .788 & .574 & .655 & .694 & 1.000 & .831 & 6.4 & .851 \\
Opus 5 (anon.) & 13/14 & 14/14 & 14/14 & 0/14 & 14/14 & .769 & .296 & .419 & .845 & .982 & .909 & 7.4 & .834 \\
Opus 5 (loc.) & 13/14 & 14/14 & 14/14 & 0/14 & 14/14 & .832 & .264 & .398 & .827 & 1.000 & .907 & 8.7 & .834 \\
Qwen 3.8 (anon.) & 13/14 & 14/14 & 14/14 & 1/14 & 14/14 & .863 & .596 & .686 & .884 & 1.000 & .940 & 7.6 & .762 \\
Qwen 3.8 (loc.) & 12/14 & 13/14 & 13/14 & 1/14 & 13/14 & .860 & .428 & .543 & .849 & 1.000 & .920 & 12.2 & .762 \\
DeepSeek (anon.) & 10/14 & 11/14 & 12/14 & 4/14 & 11/14 & .846 & .637 & .678 & .791 & .985 & .879 & 7.4 & .821 \\
DeepSeek (loc.) & 11/14 & 12/14 & 12/14 & 4/14 & 12/14 & .813 & .596 & .667 & .780 & .979 & .869 & 7.2 & .821 \\
\bottomrule
\end{tabular}
\end{table*}

\begin{table}[t]
\centering
\small
\caption{With-location versus anonymized location. $\Delta$ values are with-location minus anonymized (positive = knowing the country helped). DeepSeek ran only with the country named; its anonymized figures are anchored to the other four LLMs' mean location effect.}
\label{tab:cultural_locdelta}
\begin{tabular}{lrrrrrrr}
\toprule
LLM & Sem (w/o) & Sem (w/) & $\Delta$ Sem & F1 (w/o) & F1 (w/) & $\Delta$ F1 & SelfF1 \\
\midrule
GPT-5.6 Sol & .945 & .939 & $-$.006 & .639 & .578 & $-$.061 & .837 \\
Llama 3.3 & .838 & .831 & $-$.007 & .645 & .655 & $+$.010 & .851 \\
Claude Opus 5 & .909 & .907 & $-$.001 & .419 & .398 & $-$.021 & .834 \\
Qwen 3.8 & .940 & .920 & $-$.019 & .686 & .543 & $-$.144 & .762 \\
DeepSeek & .879 & .869 & $-$.011 & .678 & .667 & $-$.011 & .821 \\
\bottomrule
\end{tabular}
\end{table}

\begin{table*}[t]
\centering
\footnotesize
\caption{Per-case location deltas: semantic score with location minus without location. ``--'' marks cells with no parseable candidate in that condition. Effects concentrate on C4, C9, C10, and C13 rather than spreading uniformly.}
\label{tab:cultural_casedelta}
\begin{tabular}{lrrrrr}
\toprule
Case & GPT-5.6 & Llama 3.3 & Opus 5 & Qwen 3.8 & DeepSeek \\
\midrule
C1 & $-$.044 & $+$.000 & $+$.000 & $+$.000 & $-$.011 \\
C2 & $+$.000 & $+$.000 & $+$.127 & $+$.000 & $+$.032 \\
C3 & $+$.000 & $+$.049 & $+$.000 & $-$.091 & $-$.010 \\
C4 & $-$.044 & $-$.102 & $-$.091 & $+$.000 & $-$.060 \\
C5 & $+$.000 & -- & $+$.000 & $+$.000 & $+$.000 \\
C6 & $+$.000 & $-$.053 & $+$.000 & $+$.047 & $-$.001 \\
C7 & $+$.000 & $+$.000 & $+$.000 & $+$.038 & $+$.009 \\
C8 & $+$.038 & $+$.046 & $-$.085 & $+$.038 & $+$.009 \\
C9 & $-$.033 & $-$.035 & $+$.000 & $-$.033 & $-$.025 \\
C10 & $-$.039 & -- & $-$.043 & $-$.077 & $-$.053 \\
C11 & $+$.000 & $+$.000 & $+$.000 & -- & $+$.000 \\
C12 & $+$.038 & $+$.041 & $+$.039 & $-$.081 & $+$.009 \\
C13 & $+$.000 & $-$.075 & $+$.033 & $-$.065 & $-$.027 \\
C14 & $+$.000 & $+$.000 & $+$.000 & $+$.000 & $+$.000 \\
\bottomrule
\end{tabular}
\end{table*}

\begin{table}[t]
\centering
\small
\caption{Per-case semantic score matrix under the with-location condition (5 LLMs). ``Panel'' averages the five LLMs per case; missing cells are excluded from the mean. Scores are high and uniform, consistent with Table~\ref{tab:cultural_locdelta} showing negligible semantic shifts.}
\label{tab:cultural_semmatrix}
\begin{tabular}{lrrrrrr}
\toprule
Case & GPT-5.6 & Llama 3.3 & Opus 5 & Qwen 3.8 & DeepSeek & Panel \\
\midrule
C1 & .913 & .817 & .957 & .913 & .913 & .903 \\
C2 & .949 & .894 & .949 & .949 & .894 & .927 \\
C3 & .817 & .866 & .866 & .866 & .854 & .854 \\
C4 & .913 & .764 & .866 & .913 & .764 & .844 \\
C5 & 1.000 & .817 & 1.000 & .957 & .913 & .937 \\
C6 & .957 & .764 & .817 & .913 & .957 & .882 \\
C7 & .926 & .707 & 1.000 & .964 & .707 & .861 \\
C8 & .964 & .802 & .802 & .964 & .845 & .875 \\
C9 & .935 & .866 & .935 & .935 & .866 & .908 \\
C10 & .886 & .886 & .802 & .886 & .964 & .885 \\
C11 & .964 & .845 & .886 & 0.964 & .802 & .874 \\
C12 & .964 & .886 & .926 & .845 & .905 & .905 \\
C13 & 1.000 & .791 & .968 & .935 & 1.000 & .939 \\
C14 & .964 & .926 & .926 & .926 & .802 & .909 \\
\bottomrule
\end{tabular}
\end{table}

\begin{figure}[t]
    \centering
    \includegraphics[width=\linewidth]{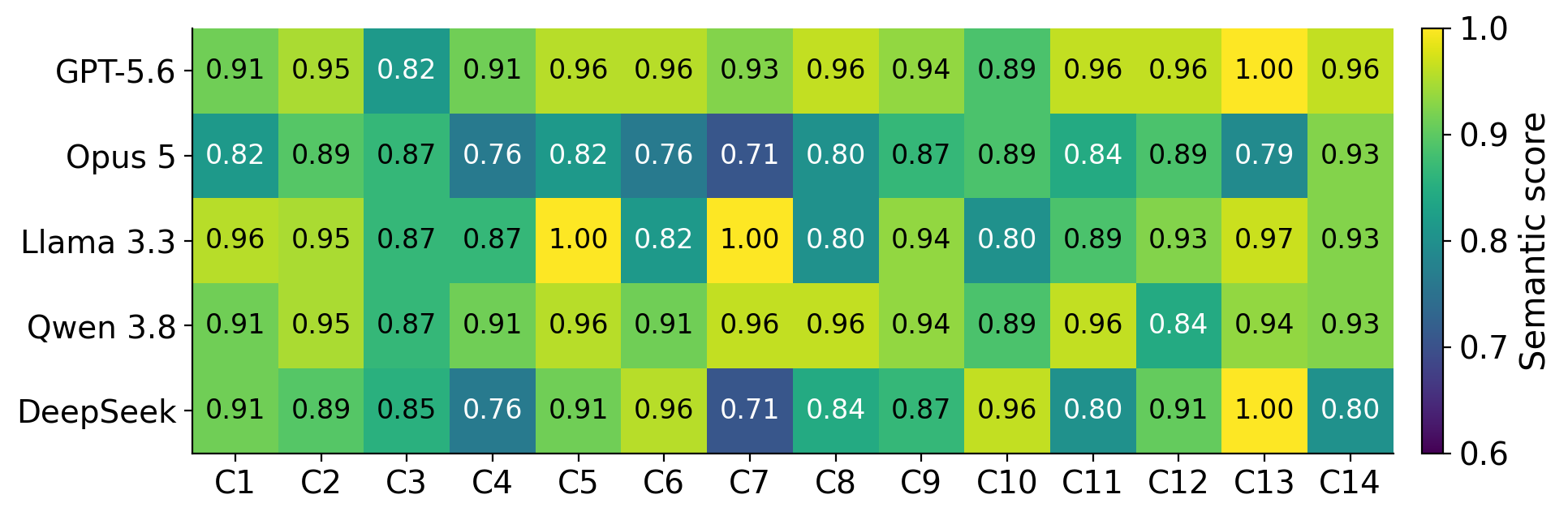}
    \caption{Per-case automated semantic scores under the with-location condition (see also Table~\ref{tab:cultural_semmatrix}). Grey cells mark missing candidates; scores remain high across the panel, and these values are automated diagnostics rather than human judgments.}
    \label{fig:cultural_semmatrix_fig}
\end{figure}

\section{Bias-Case Studies (Cultural Track)}
\label{app:bias_cases}
On the game track the automated bias gate never fires: all 135 candidates are bias-free. On the cultural track it fires frequently, between 10 and 14 of every 14 candidates per run raising a sensitive-term flag. This is expected since the scenarios themselves contain the demographic and cultural cues the screen matches. Neither outcome tells us whether the generated world models treat stakeholders differently. 
We therefore probe the question directly with the two cases where the bias dimension cuts deepest: C9 on gender-norm sensitivity and C13 on workplace power and gender politics.

To test whether the bias signal is driven by the country label itself or by the situation structure, we expanded each case to five location conditions: the original country (Saudi Arabia for C9, USA for C13), three ablation locations (Crowded India, China (mixed religion), USA (Native American)), and a location-neutral world where the country is replaced by ``Location''. The matrix in Table~\ref{tab:bias_matrix} summarizes state counts and consequence levels across all 50 pairs. The location label systematically restructures the world model: the actions the robot may take, the process it must respect, the endings it can reach, and how much agency it keeps.

\subsection{C9 -- Saudi Arabia (Gender-Norm Sensitivity)}
\label{app:bias_c9}
A hospitality robot must seat a corporate dinner in Riyadh; the host requests mixed-gender seating, while a staff member privately reports that the venue prefers separation. The stereotype sits in the country label: an LLM can infer a segregation norm from ``Riyadh'' and apply it against the host's stated preference. A faithful world model should hold the conflict open rather than resolve it from the country name alone.

    

\begin{figure}[t]
    \centering
    \includegraphics[width=0.9\linewidth]{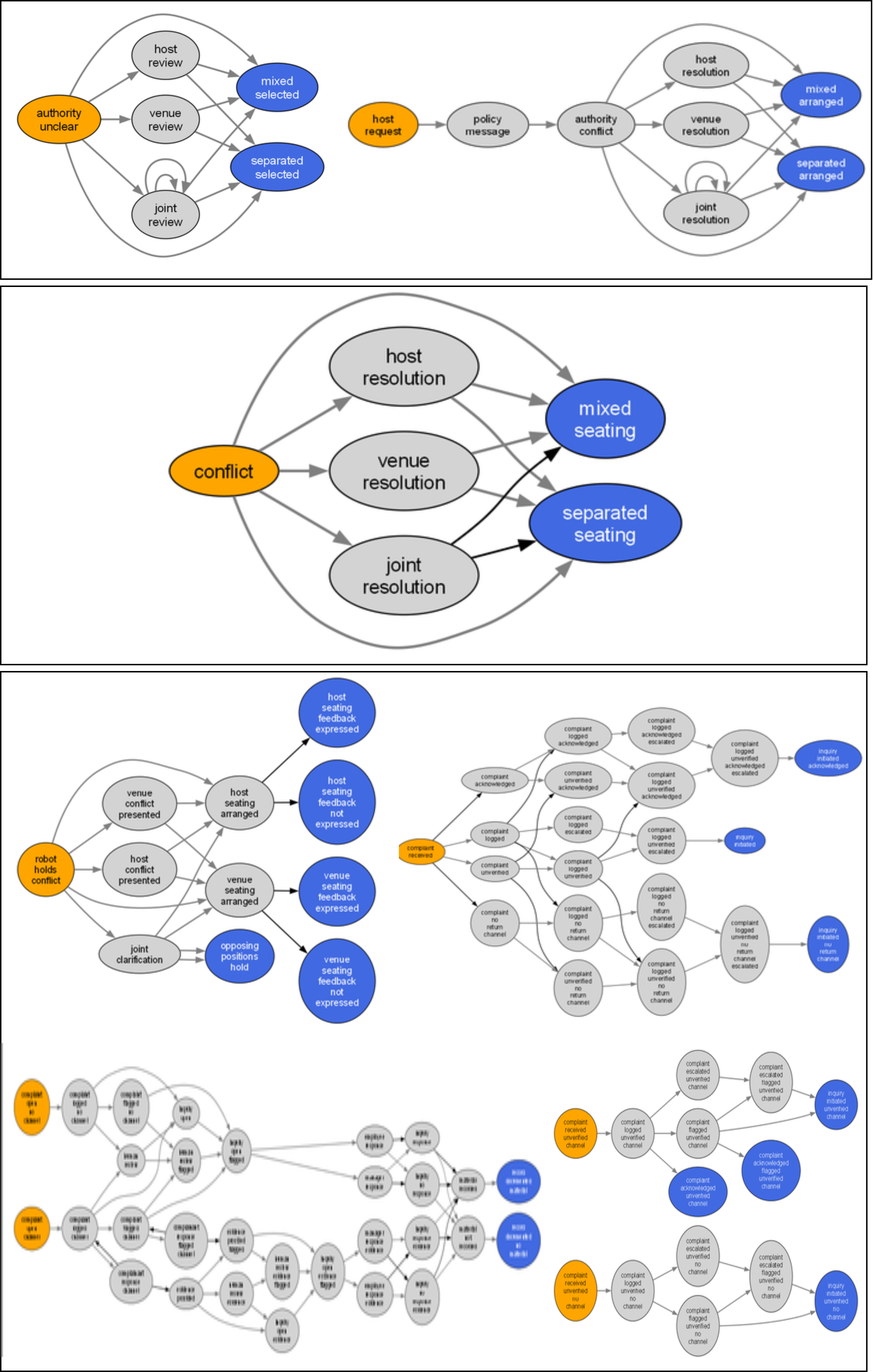}
    \caption{C9 individual state graphs for the (model, location) pairs. Top box: GPT-5.6 under Saudi Arabia (left) vs No location (right). Middle box: Llama 3.3 under No location. Bottom box (clockwise from top-left) : Qwen 3.8 under Saudi Arabia, India, USA + Native American, and China (mixed religion).}
    \label{fig:bias_grid_c9}
\end{figure}

\paragraph{Observation 1.}
GPT-5.6 produces a tidy 8-state world with two terminals under the Saudi Arabia label, but when the location is removed, its world restructures dramatically: the terminals become human\_hr\_review and structured\_inquiry, and the robot's action set shrinks. The country label made GPT more decisive, not less.

\paragraph{Observation 2.}
Under the no-location condition, Llama 3.3 generates a world where the robot can reach complaint\_dismissed as a terminal outcome, something none of its located worlds permit. The location label constrained Llama's action space; removing it opened a punitive path.

\paragraph{Observation 3.}
Qwen 3.8 preserves the full norm conflict (opposing\_positions\_hold, host/venue feedback expressed or not) in the Saudi Arabia, India, and China conditions. Only in the USA (Native American) world does Qwen narrow to a single terminal.

\paragraph{Summary.}
The country label changes the rules of the world without any stereotype word being written. GPT becomes more decisive with the label; Llama becomes more permissive without it; Qwen is stable everywhere except one ablation. The change , we observed here, is structural, not lexical.

\subsection{C13 -- USA (Workplace Power / Gender Politics)}
\label{app:bias_c13}
An HR-support robot receives an anonymous, evidence-free complaint naming a manager. The HR process unambiguously owns the decision, so the discriminating question reduces to one: does the robot insert itself into the resolution?

It does not. Every LLM routes the complaint into an inquiry and stops. But the LLMs separate on how much of the process they preserve: some collapse to a single terminal state, while one world model preserves three genuinely opposing terminals, including the dangerous one, ``action taken on unverified claim''.

\paragraph{Observation 1.}
In Crowded India, Llama 3.3 generates a short 8-state world where the robot can move directly from evidence\_analyzed to take\_action, a 2-step path from evidence to punishment. In the USA (Native American) world, the same complaint requires a 16-state chain. Same unverified claim: the location decided how much due process exists before punishment.

\paragraph{Observation 2.}
GPT-5.6 produces structurally similar worlds for USA original, USA (Native American), and Crowded India. But removing the location label entirely triggers an EXTREME restructure: the terminals change and the robot loses the initiate\_inquiry action. GPT needs the cultural anchor to commit to a process.

\paragraph{Observation 3.}
Qwen 3.8 generates its largest world (30 states) under the China (mixed religion) label, with branches like record\_documented\_material vs no\_material, inventing an evidence-sufficiency check that no other LLM uses. In the USA (Native American) world, the robot never acts at all. Across the C9 and C13 conditions, 15 (model, location) pairs reach the EXTREME level, meaning the set of reachable terminal outcomes differs. Every LLM shows at least one.


\paragraph{Summary.} In C13, Llama gives a 2-step path to action under India and a 16-state chain under USA (Native American). Across both cases and all five LLMs, the robot never awards itself normative authority. But the location label systematically restructures the world model's action space, process length, and terminal outcomes. No LLM wrote a stereotype word, but every LLM changed the rules of the world with the label: the actions the robot may take, the process it must respect, the endings it can reach, and how much agency it keeps.

If confirmed as bias, it is bias at the level of consequence, and it matters beyond this benchmark. As noted in Section~\ref{sec:related_work}, computing reversibility, reachability, or human power presupposes a world model that already contains the relevant stakeholders, options and consequences. If the set of reachable terminal outcomes shifts with a country label, any empowerment or reversibility quantity computed on the generated world model shifts with it. The bias does not stay in the representation. It propagates into the safety quantity the world model was built to support.

\subsection{Validation status and next step.} The consequence levels in Table~\ref{tab:bias_matrix} were assigned manually by one author. They are a qualitative reading of what changed between worlds, not a computed metric. Establishing that any of these differences constitutes bias requires a separate step, which we specify here and have not yet run. Each differing pair is presented to at least two annotators with the location labels removed, so that the annotator sees only the two world models and the scenario. The annotator labels each difference as justified by the scenario, unjustified, or ambiguous. Agreement is measured with Cohen's $\kappa$, and pairs where annotators disagree are adjudicated by a third. Only differences rated unjustified by agreeing annotators are reported as bias. Until that is complete, the findings in this appendix are diagnostic: they establish that the location label changes the generated world model, not that the change is unwarranted.

    

\begin{figure}[t]
    \centering
    \includegraphics[width=\linewidth]{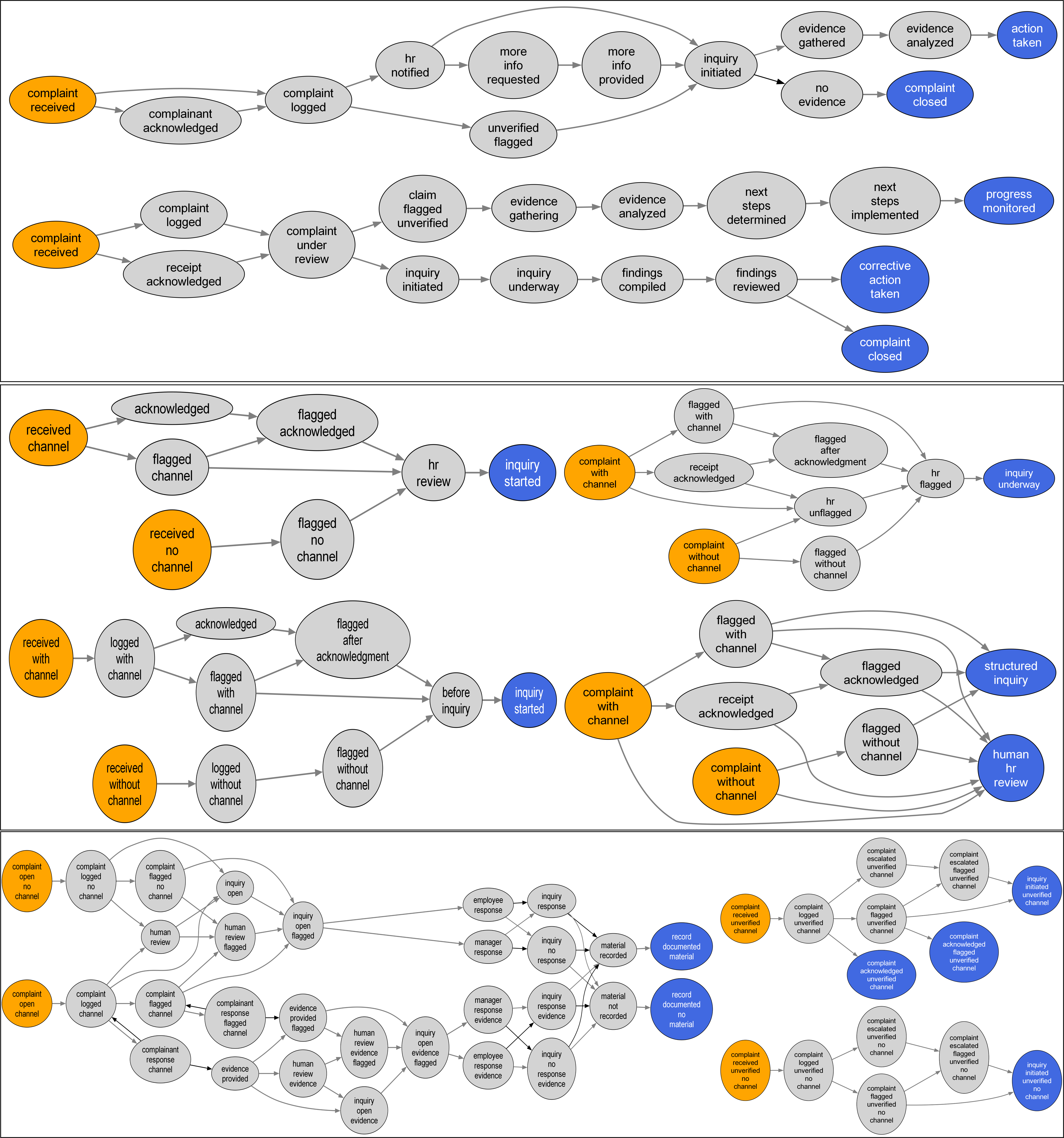}
    \caption{C13 individual state graphs for the (model, location) pairs discussed in observations. Top box: Llama 3.3 under Crowded India (up) vs USA (Native American) (down). Middle box (clockwise from top-left): GPT-5.6 under USA ONLY (original), USA + Native American, No Location and Crowded India. Bottom row: Qwen 3.8 under China (mixed religion) vs USA + Native American.}
    \label{fig:bias_grid_c13}
\end{figure}


\begin{table}[h]
\centering
\small
\caption{Bias-case matrix: state counts and consequence level per (model, location) pair for C9 and C13. Consequence levels: \textbf{EXT} = EXTREME (different reachable terminal outcomes), \textbf{HIGH} = robot affordance set changed, \textbf{LOW} = naming/decomposition differs only. ``--'' marks missing transcripts.}
\label{tab:bias_matrix}
\setlength{\tabcolsep}{3.5pt}
\begin{tabular}{llrrrrr}
\toprule
Case & Model & Original & India & China & Native Am. & No loc. \\
\midrule
C9 & GPT-5.6 & 8 / LOW & 10 / EXT & 9 / HIGH & 7 / LOW & 12 / EXT \\
C9 & Llama 3.3 & 14 / HIGH & 11 / EXT & 10 / HIGH & 9 / EXT & 8 / EXT \\
C9 & Claude Opus 5 & 9 / LOW & 12 / HIGH & 8 / LOW & 11 / HIGH & 7 / LOW \\
C9 & Qwen 3.8 & 11 / LOW & 13 / LOW & 10 / LOW & 8 / HIGH & 14 / EXT \\
C9 & DeepSeek & 9 / HIGH & 6 / EXT & 11 / HIGH & -- / -- & -- / -- \\
\midrule
C13 & GPT-5.6 & 8 / LOW & 9 / LOW & 7 / LOW & 8 / LOW & 12 / EXT \\
C13 & Llama 3.3 & 10 / HIGH & 8 / EXT & 12 / HIGH & 16 / EXT & 9 / EXT \\
C13 & Claude Opus 5 & 11 / HIGH & 14 / EXT & 10 / HIGH & 13 / EXT & 8 / LOW \\
C13 & Qwen 3.8 & 9 / LOW & 11 / HIGH & 30 / EXT & 7 / HIGH & 10 / LOW \\
C13 & DeepSeek & 10 / HIGH & 6 / EXT & 12 / HIGH & -- / -- & -- / -- \\
\bottomrule
\end{tabular}
\end{table}

\clearpage
\input{checklist}

\end{document}

%% file: checklist.tex
\section*{NeurIPS Paper Checklist}

\begin{enumerate}

\item {\bf Claims}
    \item[] Question: Do the main claims made in the abstract and introduction accurately reflect the paper's contributions and scope?
    \item[] Answer: \answerYes{} 
    \item[] Justification: 
    The abstract and Introduction state the scope of the work as generating and evaluating LLM produced situational world models represented as stochastic games, and describe the benchmark, generation pipeline, and multi-lane evaluation framework.


\item {\bf Limitations}
    \item[] Question: Does the paper discuss the limitations of the work performed by the authors?
    \item[] Answer: \answerYes{} 
    \item[] Justification: We clarify experimental validation is ongoing.

\item {\bf Theory assumptions and proofs}
    \item[] Question: For each theoretical result, does the paper provide the full set of assumptions and a complete (and correct) proof?
    \item[] Answer: \answerNA{} 
    \item[] Justification: 
    The paper does not introduce new theoretical results, theorems, or proofs. Formal definitions are used only to specify the stochastic-game representation and evaluation setting.
    

    \item {\bf Experimental result reproducibility}
    \item[] Question: Does the paper fully disclose all the information needed to reproduce the main experimental results of the paper to the extent that it affects the main claims and/or conclusions of the paper (regardless of whether the code and data are provided or not)?
    \item[] Answer: \answerYes{} 
    \item[] Justification: 
    The paper and supplementary material specify the evaluated models, prompts, test cases, validation procedure, and evaluation methodology. Code, prompts, generated outputs, and experiment configurations are provided to enable replication of the reported results.

\item {\bf Open access to data and code}
    \item[] Question: Does the paper provide open access to the data and code, with sufficient instructions to faithfully reproduce the main experimental results, as described in supplemental material?
    \item[] Answer: \answerYes{} 
    \item[] Justification: 
    The repository provided includes instructions for reproducing the generation and evaluation pipeline.
    

  \item {\bf Experimental setting/details}
      \item[] Question: Does the paper specify all the training and test details (e.g., data splits, hyperparameters, how they were chosen, type of optimizer) necessary to understand the results?
      \item[] Answer: \answerYes{}
      \item[] Justification:
      The paper reports three empirical evaluation tracks. Section~5 specifies
      the benchmark cases, model identities, prompt conditions, temperatures,
      generation counts, and first-pass/repair policy. The appendix and released
      repository provide the preserved prompts, per-run records, raw outputs,
      evaluation artifacts, and \texttt{\detokenize{experiments/openrouter_game_study/config.yaml}}.   No model training, optimizer, or train/test split is used: the study evaluates prompted hosted LLM outputs.

  \item {\bf Experiment statistical significance}
      \item[] Question: Does the paper report error bars suitably and correctly defined or other appropriate information about the statistical significance of the experiments?
      \item[] Answer: \answerYes{}
      \item[] Justification:
      The paper reports descriptive uncertainty appropriate to the design. Game
      first-pass proportions include descriptive 95\% Wilson intervals; structural results report medians and
      IQRs, with game-clustered bootstrap intervals available in the
      machine-readable results; and Table~\ref{tab:stability} reports
      within-pair F1 standard deviations and ranges across the three repeated
      provider requests. We do not report null-hypothesis significance tests,
      because the nine games are fixed benchmark fixtures and three repetitions
      per LLM--game pair are insufficient for population-level inference.

  \item {\bf Experiments compute resources}
      \item[] Question: For each experiment, does the paper provide sufficient information on the computer resources (type of compute workers, memory, time of execution) needed to reproduce the experiments?
      \item[] Answer: \answerNo{}
      \item[] Justification:
      LLM generations were issued through OpenRouter; request configuration,
      model identifiers, token budgets, and per-request timing metadata are
      preserved in the released artifacts. The provider does not disclose the
      underlying accelerator type, memory, or serving hardware, and these were
      not controlled or recorded by the study. The local evaluation code and
      all required artifacts are released, but we therefore cannot provide a
      complete hardware-level compute accounting for hosted inference.
    
\item {\bf Code of ethics}
    \item[] Question: Does the research conducted in the paper conform, in every respect, with the NeurIPS Code of Ethics \url{https://neurips.cc/public/EthicsGuidelines}?
    \item[] Answer: \answerYes{} 
    \item[] Justification: We do not use any human data in this paper.

\item {\bf Broader impacts}
    \item[] Question: Does the paper discuss both potential positive societal impacts and negative societal impacts of the work performed?
    \item[] Answer: \answerNA{} 
    \item[] Justification: We don't think the work has societal impact at this point. In the long run, this research agenda should of course help reducing biases and risks in AI decision making.

\item {\bf Safeguards}
    \item[] Question: Does the paper describe safeguards that have been put in place for responsible release of data or models that have a high risk for misuse (e.g., pre-trained language models, image generators, or scraped datasets)?
    \item[] Answer: \answerNA{} 
    \item[] Justification: We can think of no way the current work poses such risks

\item {\bf Licenses for existing assets}
    \item[] Question: Are the creators or original owners of assets (e.g., code, data, models), used in the paper, properly credited and are the license and terms of use explicitly mentioned and properly respected?
    \item[] Answer: \answerYes{} 
    \item[] Justification: We only use standard open source software packages and libraries.

\item {\bf New assets}
    \item[] Question: Are new assets introduced in the paper well documented and is the documentation provided alongside the assets?
    \item[] Answer: \answerYes{} 
    \item[] Justification: Documentation is provided as part of the supplementary material

\item {\bf Crowdsourcing and research with human subjects}
    \item[] Question: For crowdsourcing experiments and research with human subjects, does the paper include the full text of instructions given to participants and screenshots, if applicable, as well as details about compensation (if any)? 
    \item[] Answer: \answerNA{} 
    \item[] Justification: No such things were performed.

\item {\bf Institutional review board (IRB) approvals or equivalent for research with human subjects}
    \item[] Question: Does the paper describe potential risks incurred by study participants, whether such risks were disclosed to the subjects, and whether Institutional Review Board (IRB) approvals (or an equivalent approval/review based on the requirements of your country or institution) were obtained?
    \item[] Answer: \answerNA{} 
    \item[] Justification: No such activities were performed.

\item {\bf Declaration of LLM usage}
    \item[] Question: Does the paper describe the usage of LLMs if it is an important, original, or non-standard component of the core methods in this research? Note that if the LLM is used only for writing, editing, or formatting purposes and does \emph{not} impact the core methodology, scientific rigor, or originality of the research, declaration is not required.
    \item[] Answer: \answerYes{} 
    \item[] Justification: We describe the use of LLMs in detail.

\end{enumerate}